\documentclass[acmsmall]{acmart}
\AtBeginDocument{%
  \providecommand\BibTeX{{%
    \normalfont B\kern-0.5em{\scshape i\kern-0.25em b}\kern-0.8em\TeX}}}

\setcopyright{acmcopyright}
\copyrightyear{2023}
\acmYear{2023}
\acmDOI{10.1145/3591870}

\acmJournal{TOSEM}
\acmVolume{}
\acmNumber{}
\acmArticle{}
\acmMonth{4}

\usepackage{url}            
\usepackage{amsfonts,amsmath}       
\usepackage{nicefrac}       
\usepackage{multirow}
\usepackage{tablefootnote}
\usepackage{algorithm,algorithmic}
\usepackage[normalem]{ulem}
\usepackage{arydshln}
\usepackage{caption}
\usepackage{enumitem}
\usepackage{wrapfig}
\usepackage{microtype}
\usepackage{graphicx}
\usepackage{tcolorbox}
\usepackage{booktabs} 
\usepackage{cutwin}
\usepackage[pangram]{blindtext}
\usepackage[calc]{adjustbox}
\usepackage{subcaption}
\usepackage{xcolor}
\usepackage{colortbl}

\usepackage{amsthm}

\definecolor{dg}{rgb}{0,0.694,0.298}
\definecolor{purple}{rgb}{0.4,0.176,0.569}
\definecolor{royalblue}{RGB}{65,105,225}
\definecolor{colorsteps}{rgb}{0.83, 0.83, 0.83}
\definecolor{tab_red}{rgb}{1,0.76,0.71}
\definecolor{deepgreen}{rgb}{0, 0.49, 0.25}
\definecolor{rq}{HTML}{009444}

\definecolor{royalblue}{RGB}{65,105,225} 
\definecolor{lightgray}{HTML}{eeeeee}
\definecolor{tab_red}{rgb}{1,0.76,0.71}

\usepackage{hyperref}
\hypersetup{colorlinks=true}
\hypersetup{linktoc=all}
\hypersetup{citecolor=MidnightBlue}
\hypersetup{linkcolor=BrickRed}
\hypersetup{urlcolor=MidnightBlue}
\usepackage[all]{hypcap}

\usepackage[nameinlink]{cleveref}
\creflabelformat{equation}{#2\textup{#1}#3}  
\crefname{assumption}{assumption}{assumptions}

\usepackage[colorinlistoftodos]{todonotes}

\newcommand{\sys}{{PatchCensor}\xspace} 

\newcommand{\etal}{\textit{et al}.~}
\newcommand{\ie}{\textit{i}.\textit{e}.~}
\newcommand{\eg}{\textit{e}.\textit{g}.~}

\newcommand{\vs}{\textit{vs}~}

\newcommand{\cA}{{\mathcal A}}

\newcommand{\cP}{{\mathcal P}}
\newcommand{\cT}{{\mathcal T}}

\newcommand{\cX}{{\mathcal X}}
\newcommand{\cY}{{\mathcal Y}}
\newcommand{\xp}{x^{\prime}}
\newcommand{\xpp}{x^{\prime\prime}}

\newcommand{\bE}{\mathbf{E}}

\newcommand{\reals}{\mathbb{R}}

\newcommand{\cXtrust}{\cX_{\text{robust}}}
\newcommand{\accClean}{{acc}_{\text{clean}}}
\newcommand{\accCertified}{{acc}_{\text{certified}}}
\newcommand{\accInTrust}{{acc}_{\text{in-robust}}}
\newcommand{\rTrust}{r_{\text{robust}}}

\newcommand{\responserefa}[0]{}

\newcommand{\responserefb}[0]{}

\begin{document}

\title[PatchCensor: Patch Robustness Certification for Transformers via Exhaustive Testing]{PatchCensor: Patch Robustness Certification for Transformers via Exhaustive Testing}

\author{Yuheng~Huang}
\authornote{
This work is done partially while Yuheng Huang was an intern with Yuanchun Li at Institute for AI Industry Research (AIR), Tsinghua University, China
}
\affiliation{
    \institution{University of Alberta}
    \city{Edmonton}
    \state{Alberta}
    \country{Canada}
    }
\email{yuheng18@ualberta.ca}

\author{Lei~Ma}
\affiliation{
    \institution{University of Alberta}
    \city{Edmonton}
    \state{Alberta}
    \country{Canada}
    }
\affiliation{
    \institution{The University of Tokyo}
    \city{Tokyo}
    \country{Japan}
    }
\email{ma.lei@acm.org}

\author{Yuanchun~Li}
\authornote{Yuanchun Li is the corresponding author and is also affiliated with Shanghai AI Laboratory,China.}
\affiliation{
    \institution{Institute for AI Industry Research (AIR), Tsinghua University}
    \city{Beijing}
    \country{China}
    }
\email{liyuanchun@air.tsinghua.edu.cn}

\renewcommand{\shortauthors}{Huang et al.}


\begin{abstract}
    In the past few years, Transformer has been widely adopted in many domains and applications because of its impressive performance. Vision Transformer (ViT), a successful and well-known variant, attracts considerable attention from both industry and academia thanks to its record-breaking performance in various vision tasks. However, ViT is also highly nonlinear like other classical neural networks and could be easily fooled by both natural and adversarial perturbations. This limitation could pose a threat to the deployment of ViT in the real industrial environment, especially in safety-critical scenarios. How to improve the robustness of ViT is thus an urgent issue that needs to be addressed. Among all kinds of robustness, patch robustness is defined as giving a reliable output when a random patch in the input domain is perturbed. The perturbation could be natural corruption, such as part of the camera lens being blurred. It could also be a distribution shift, such as an object that does not exist in the training data suddenly appearing in the camera. And in the worst case, there could be a malicious adversarial patch attack that aims to fool the prediction of a machine learning model by arbitrarily modifying pixels within a restricted region of an input image. This kind of attack is also called physical attack as it is believed to be more real than digital attack. Although there has been some work on patch robustness improvement of Convolutional Neural Network (CNN), related studies on its counterpart ViT are still at an early stage as ViT is usually much more complex with far more parameters. It is harder to assess and improve its robustness, not to mention to provide a provable guarantee.
    
    In this work, we propose \sys, aiming to certify the patch robustness of ViT by applying exhaustive testing. We try to provide a provable guarantee by considering the worst patch attack scenarios. Unlike empirical defenses against adversarial patches that may be adaptively breached, certified robust approaches can provide a certified accuracy against arbitrary attacks under certain conditions. However, existing robustness certifications are mostly based on robust training, which often requires substantial training efforts and the sacrifice of model performance on normal samples. To bridge the gap, \sys seeks to improve the robustness of the whole system by detecting abnormal inputs instead of training a robust model and asking it to give reliable results for every input, which may inevitably compromise accuracy. Specifically, each input is tested by voting over multiple inferences with different mutated attention masks, where at least one inference is guaranteed to exclude the abnormal patch. This can be seen as complete-coverage testing, which could provide a statistical guarantee on inference at the test time. Our comprehensive evaluation demonstrates that PatchCensor is able to achieve high certified accuracy (\eg 67.1\% on ImageNet for 2\%-pixel adversarial patches), significantly outperforming state-of-the-art techniques while achieving similar clean accuracy (81.8\% on ImageNet). The clean accuracy is the same as vanilla ViT models. Meanwhile, our technique also supports flexible configurations to handle different adversarial patch sizes by simply changing the masking strategy.
\end{abstract}



\maketitle

\section{Introduction}
\label{sec:introduction}

As the core to the success of many intelligent systems, Deep neural networks (DNNs) are widely adopted across various industrial applications and domains. With the substantial advance in Deep Learning (DL) techniques, there is also an increasing trend to deploy DL models in the physical world in safety-critical scenarios, \eg, face authentication on smartphones~\cite{sun2016sparsifying}, driving assistance on autonomous cars~\cite{wei2019enhanced}, and intruder detection on surveillance cameras~\cite{chung2017two}. However, DNNs are well-known for their nonlinear property, which makes them highly sensitive to perturbation in their input space. Both natural perturbations~\cite{hendrycks2018benchmarking} and evasion attacks~\cite{szegedy2013intriguing,papernot2018sok} could pose threats to these DL models. As an alternative to Convolutional Neural Network (CNN) in vision tasks and recurrent language model (\eg RNN and LSTM) in language processing tasks, Transformers~\cite{vaswani2017attention, dosovitskiy2021vit}, the current state-of-the-art DNNs, still suffer from this sensitivity. This limitation is one of the major threats to deploying DNNs in safety-critical scenarios. 

To address this issue, both the SE community~\cite{kim2019guiding, stocco2020misbehaviour, zhang2020towards, zhang2020uncertainty, wang2020dissector, li2020operational, zhang2020towards, ma2021test, zhou2020deepbillboard, wang2019adversarial, du2019deepstellar, zhang2020dynamic, zhao2021attack} and the AI community~\cite{naseer2019local,rao2020adversarial, chen2021turning, chou2020sentinet, rossolini2022real, chiang2020certified,xiang2020patchguard,levine2020randomized,zhang2020clipped,metzen2021bagcert, salman2021certified, mccoyd2020minority, xiang2021patchguard++, han2021scalecert} are trying to improve the  DL models' robustness through training, testing and verification. Related work can be roughly categorized into two types: (1) Improve model robustness in the digital domain. (2) Improve model robustness in the physical domain. The issue of digital domain robustness was first discovered by Goodfellow~\etal, and it is formalized as the well-known adversarial attack~\cite{goodfellow2014explaining}. The model is defined as robust when its prediction is stable with a adversarially generated \textit{global} $L_p$-bounded perturbation ($p = 1, 2, \infty, ...$). 

{\responserefa{}
More recently, researchers have found that the attacks applied to the whole image with small perturbations are hard to implement in real world. The reasons are four folds: (1) In practical scenarios, noisy physical environments can destroy the perturbations created by digital attack~\cite{lu2017no}; (2) The small magnitude perturbation may be missed by the camera because of sensor imperfections~\cite{evtimov2017robust}. (3) It is challenging to create an attack modifying the whole background in the real world~\cite{evtimov2017robust}. (4) Printing small magnitude perturbations is less effective~\cite{evtimov2017robust, zhou2020deepbillboard}. Because of these practical concerns, patch attack (or physical attack)~\cite{brown2017adversarial, zhou2020deepbillboard, evtimov2017robust, song2018physical, thys2019fooling} is proposed, which is believed to be a real threat and has a more realistic attacking scenario than digital attacks. Its perturbation is limited in a given region with no magnitude restriction. Along the line, researchers found that an attacker can fool the DNN in a self-driving car by attaching a printed sticker onto the road signs~\cite{eykholt2018robust} or bypass face authentication by presenting a customized object in front of the camera~\cite{komkov2021advhat}. In general, patch attack can be very helpful for the developer to understand the vulnerability of the model's robustness under the patch-liked attacks/noises, so that to further propose techniques to counteract since they can be highly relevant and linked to the safety and security issues. However, as the patch attack has a totally different attack model from the digital attack, the defense for the digital attack is usually hard to apply. With this real threat, it is thus urgent and necessary to propose a defense mechanism that is specifically designed for the patch attack.

}




{\responserefa{}
    With the advance of related study of the model's robustness against adversarial distribution shift, there is also an ongoing trade in studying the natural distributional robustness of DNNs. Its motivation is that the worst-case adversarial perturbations are unlikely to happen in reality, while the corruptions usually seen in the natural environment can seriously influence models' performance. Similar to the adversarial setting, natural corruption can also be applied to the whole image or a limited patch. Examples of the former can be gaussian noise or motion blur of cameras~\cite{hendrycks2018benchmarking} in the entire image. For the latter, it can be noise restricted to a patch~\cite{gu2022evaluating} or occlusion in object detection~\cite{zhang2018deepvoting} and anomaly objects in the open-set video setting~\cite{acsintoae2022ubnormal}. 

    Regarding digital-domain robustness, the adversarial perturbations and natural perturbations usually cannot be analyzed in the same framework as the attack magnitude is restricted by $L_p$-bounded while the natural perturbations are generally unbounded. A recent study also points out that there is a tradeoff between adversarial and natural robustness~\cite{moayeri2022explicit}. However, the patch robustness of both adversarial and natural perturbations can actually be analyzed together. Both set no restriction for perturbation magnitude, and only the patch size is limited. In this sense, given the same patch region, if the robustness of the worst-case adversarial attack is guaranteed, then the robustness of natural perturbations is also ensured. 
}


Although patch robustness is important as it is closely related to both adversarial and natural threats, related studies are still at the early stage. Various heuristic defenses~\cite{naseer2019local,rao2020adversarial} have been designed to improve patch robustness. However, these heuristic approaches are unable to guarantee robustness against unseen adaptive attacks~\cite{chiang2020certified}. Some provable defenses were recently proposed for adversarial patches. Most of them \cite{zhang2020clipped,levine2020randomized,xiang2020patchguard,metzen2021bagcert} aim to design a model that can guarantee consistent predictions on images with or without an adversarial patch. These approaches are mostly based on the insight that the prediction can be made by aggregating local features extracted from small independent receptive fields. The certified robustness is achieved by ensuring that the aggregated result will not be dominated by a small number of potentially adversarial receptive fields. As a result, they have to significantly compromise the prediction accuracy due to the lack of global features. One recent work~\cite{xiang2021patchguard} in this type can only achieve a certified accuracy of 26.0\% on ImageNet under 2\%-pixel adversarial patches.

Instead of attempting to hazardously make imprecise predictions in the presence of abnormal patches, we argue that it might be more reasonable to warn the user if the input is potentially harmful. For example, imagine a DNN is used for driving assistance, it would be desirable for the model to show how reliable and precise it is when it gives a driving suggestion while informing the driver about the risk if there is a situation that the model can not handle.
Certified detection is designed for this purpose.
The Minority Reports Defense~\cite{mccoyd2020minority}, PatchGuard++~\cite{xiang2021patchguard++}, and ScaleCert~\cite{han2021scalecert} all seek to detect the presence of an adversarial patch by partially occluding the image around each candidate patch location and analyzing the predictions of all occluded images. However, the Minority Reports Defense cannot scale to complex high-resolution images (\eg ImageNet) due to the training difficulty and heavy computation overhead, and PatchGuard++ and ScaleCert are based on CNN backbones with small receptive fields (\eg BagNet), which cannot sufficiently utilize the global feature.
As a result, how to obtain accurate and scalable robustness certification against adversarial patches remains an open problem.

To this end, we introduce \sys, a testing-based method to certify the robustness against adversarial patches that can flexibly and accurately scale to complex images and different adversarial patch sizes. \sys falls in the category of certified detection and aims to warn the user whenever an abnormal patch appears. As a testing-based method \sys is zero-shot (aka. requires no additional training) by directly reusing a pretrained ViT backbone for certification. By providing the provable guarantee for the worst-case patch attacks, our method could also be applied to the detection of the corrupted natural patch. 

\sys is explicitly designed for Transformer architecture because:

\begin{itemize}[leftmargin=*]
    \item{Transformers have achieved state-of-the-art performance across both Natural Language Process (NLP) and Computer Vision (CV) tasks.}
    
    \item{Recent studies show that Transformers are more vulnerable against physical-domain adversarial patches~\cite{gu2021vision}, addressing the need to defend it.}
    
    \item{Prior work has shown that Transformer-based neural network architecture is robust against absent blocks~\cite{paul2021vision, naseer2021intriguing}, making it more reasonable to design mutation rules that explicitly mask some blocks.}
\end{itemize}

Although our mutation strategy is designed for Transformers and works at the abstract feature block level, we will focus on Vision Transformer (ViT) in the rest of the paper because: (1) Most related work is designed and evaluated on vision tasks, and it would be more reasonable to compare with them in the same domain. (2) Both adversarial patch attacks and corresponding defense are formally defined in vision tasks but not in the natural language domain. 

In a ViT model, the input image is partitioned into small patches. The patches are fed into several Transformer encoder layers to exchange local information by attending to each other. The global feature is obtained after the Transformer layers and used to classify the whole image. Such an attention mechanism enables a convenient way to exclude local patches from an inference pass - a patch can be excluded by masking the attention of other patches towards it. The remaining patches can still interact with each other to obtain sufficient global information for prediction.

Given an input image of a machine learning model, \sys checks whether the prediction of the model may be controlled by an adversarial patch by testing it with different mutations.
In each mutation, we exclude a region of the image by manipulating the attention mask of ViT and check whether the mutated inference can produce the same result as the original prediction.
The exhaustive testing is done if all possible locations of the adversarial patch are tested, and an input is certifiably robust if all the tests reach a consensus on the same prediction.
By designing the optimal strategy of attention mask generation, \sys can efficiently achieve full test coverage with a minimal number of mutations.
Given a data distribution, the robustness certification of \sys is obtained by running exhaustive testing on the samples in the distribution. Similar to most certification approaches~\cite{lecuyer2019certified, wong2018provable}, the certified accuracy is calculated as the ratio of samples that are correctly predicted and certifiably robust.

We evaluate \sys on popular datasets, including CIFAR-10, GTSRB, Food-101 and ImageNet, with adversarial patch sizes ranging from 0.4\% to 25\%.
The results demonstrate that \sys is able to achieve significantly higher clean accuracy and certified accuracy than existing certified defenses.
In particular, \sys is able to stably achieve a high test accuracy in the verified image subset under different adversarial patch sizes, although the ratio of verifiable inputs decreases under larger adversarial patches.

Finally, we discuss the advantages and disadvantages of \sys over the traditional certified recovery approaches to assess whether we have effectively addressed the limitations identified in the preliminary study. We empirically demonstrate the superior ability of \sys in dealing with images with a small {\responserefa{}region of interest (ROI)}. We also highlight the difference between the statistical certificate obtained by \sys and traditional certificates, which has rarely been discussed in literature before.

In summary, this paper makes the following contributions:

\begin{itemize}[leftmargin=*]

    \item We originally propose a zero-shot, testing-driven certified defense against adversarial patches by utilizing the characteristics of Transformer-based DNN architecture.
    \item We implement and test our method on popular datasets, including CIFAR-10, GTSRB, Food-101 and ImageNet, and we demonstrate that our method can achieve state-of-the-art certified accuracy on the datasets.
    \item We discuss the fundamental differences between certified recovery and certified detection through experiments, which explains the better practicality of our approach. 
\end{itemize}

To the best of our knowledge, \sys is a very early work that investigates the machine learning software security certification from the combined view of Transformer and exhaustive testing. We demonstrate that this could be a promising direction with the advantage and practical usefulness compared with other state-of-the-art techniques, which potentially inspires future research studies to safeguard the security of data-driven intelligent software.

\section{Background and Related Work}
\label{sec:background}

In this section, we first briefly introduce the background of the Vision Transformer (ViT) and related work about its robustness. Then we discuss previous work that is relevant to safeguarding DNNs in two directions: (1) DNN testing, which aims to improve the safety, quality and robustness of DL models through empirical testing; (2) Patch Attack defense, which aims to defend DL models against the worst-case adversarial threat. 

\subsection{Vision Transformer and Its Robustness}


The Vision Transformer (ViT) architecture~\cite{dosovitskiy2021vit} is proposed recently as an alternative to Convolution Neural Networks (CNNs for short) \cite{szegedy2015going} in computer vision tasks.
ViT is inspired by the scaling success of self-attention-based architectures (in particular Transformer~\cite{vaswani2017attention}) on NLP tasks. In a ViT, an image is split into patches and converted to a sequence of linear embeddings of these patches, which is then fed into a Transformer as the input. Image patches are treated the same way as tokens (words) in an NLP application. By pre-training the model on large amounts of data, it is able to attain excellent results as compared to convolutional networks.
Later studies \cite{liu2021swin,zhao2021point,zhang2022dino} have shown that the Transformer architecture can achieve state-of-the-art performance on most computer vision (CV) tasks.

Despite the great success of ViT, its robustness against adversarial attacks is still in doubt. Researchers have found that ViT is robust towards occlusion and natural perturbations~\cite{naseer2021intriguing, bhojanapalli2021understanding, paul2021vision}, while it is vulnerable to adversarial attacks~\cite{bai2021transformers, gu2021vision, mahmood2021robustness, fu2021patch}. Bai~\etal~\cite{bai2021transformers} claim that transformers are at least no more robust than CNNs when defending digital-domain adversarial perturbation, while Gu~\etal~\cite{gu2021vision} point out that ViT is significantly more vulnerable against physical-domain adversarial patches. Furthermore, Fu~\etal~\cite{fu2021patch}. develop an attack framework called Patch-Fool that targets the self-attention mechanism of the ViT specifically. They also find that ViTs become weaker learners compared to CNNs under this setting, addressing the need for improving the robustness of ViT.
In this work, we aim to design a certified adversarial patch defense based on ViT by utilizing its strength (robustness to occlusion) to mitigate its weakness (vulnerability against adversarial attacks).

\subsection{DNN Testing}

How to improve the quality and robustness of DNNs and ensure their correctness at the deployment stage remains to be an open problem. In recent years we have witnessed ongoing efforts in the SE community trying to do quality assurance for DL models through testing. They can be roughly categorized into two types: (1) those that aim to assess the quality of DNNs before deployment by synthesizing test inputs and (2) those that aim to capture subtle behaviours exhibited in DL models at a single input level. 

The former is motivated by the fact that testing data are expensive and often insufficient to assess the quality of DNNs. Motivated by this, some testing techniques are developed to synthesize test inputs to thoroughly test the DNNs~\cite{ma2019deepct, ma2018deepmutation, xie2019deephunter, tian2018deeptest, zhang2018deeproad, zhou2020deepbillboard, wang2020metamorphic}. The test generation process can be guided through a series of DNN coverage criteria~\cite{pei2017deepxplore, ma2018deepgauge}. The most relevant one is DeepBillBoard~\cite{zhou2020deepbillboard} which aims to generate patch perturbations to test Autonomous Driving systems. However, these testing techniques are parallel to our work as they focus on evaluating the quality of DNNs at the development stage while we try to detect abnormal situations given a single input at the deployment stage. 

Another line of work tries to probe the boundaries of DNNs' capability with a single test input. Abnormal behaviour analysis tries to analyze the models' behavior and understand when the model could yield erroneous prediction~\cite{kim2019guiding, stocco2020misbehaviour, zhang2020towards, zhang2020uncertainty, wang2020dissector, li2020operational, ma2021test}. This analysis can be done by assessing the models' uncertainty~\cite{zhang2020towards, zhang2020uncertainty, ma2021test}, by detecting whether the input instance is Out-of-Distribution (OOD) or not~\cite{kim2019guiding, wang2020dissector, li2020operational}, or by directly predicting the failure of the AI system~\cite{li2020operational}. Adversarial detection~\cite{wang2019adversarial, du2019deepstellar, zhang2020dynamic, zhao2021attack} has a slightly different focus and tries to detect inputs that are adversarially tampered with. Our work is orthogonal to them as we aim to perform exhaustive testing for a given input to achieve a certified level guarantee, while most related work attempts to provide empirical results.

Our work is built on a specific technique called {\responserefa{} testing by mutation}. It has been widely used in software engineering and has recently been adapted for testing complex DL-driven systems in both the development~\cite{ma2018deepmutation, zhang2018deeproad, tian2018deeptest, xie2019deephunter, riccio2021deepmetis, humbatova2021deepcrime, liu2021dialtest, huang2021coverage} and deployment~\cite{wang2019adversarial} stages. The key to this technique is to design suitable mutation operators carefully as different tasks (\eg CV~\cite{zhang2018deeproad, tian2018deeptest, xie2019deephunter, riccio2021deepmetis} or NLP~\cite{huang2021coverage, liu2021dialtest}) and different targets (\eg model input mutation~\cite{zhang2018deeproad, tian2018deeptest, xie2019deephunter, riccio2021deepmetis, huang2021coverage, liu2021dialtest} or model mutation~\cite{ma2018deepmutation, humbatova2021deepcrime}) may require totally different operators. \sys apply exhaustive testing for mutated model inputs at the deployment stage. The mutation operator is designed at the abstract feature level. Although our discussion mostly centers around the context of computer vision, our proposed mutation technique is general and could also potentially be applied to other application scenarios with some adaptation.

\subsection{Adversarial Patch Attack and Defense}

Many studies have shown that deep neural networks (DNNs) could be vulnerable against adversarial attacks~\cite{szegedy2013intriguing,goodfellow2014explaining}.
Classical adversarial attacks are known as adversarial perturbations, aiming to generate a small perturbation added to the pixels of an image that can lead to misclassification.

Instead of the whole-image perturbations of classical adversarial attacks that may be unrealistic in the real world \cite{chiang2020certified}, adversarial patch attack performs pixel modification within a restricted region in the image, which is easily achievable in the physical environment (\eg, putting a sticker on the traffic sign, wearing a specially designed glass or T-shirt, etc.). Brown~\etal \cite{brown2017adversarial} first demonstrated the feasibility of fooling an image classifier with a specifically crafted physical patch.
Along this direction, numerous other approaches \cite{karmon2018lavan,eykholt2018robust,kong2020physgan, zhou2020deepbillboard, sato2021dirty} have been proposed to achieve more practical and effective patch attacks, or test real-world AI-based software systems under the adversarial patches.


To better ensure the reliability of AI-based software, there also comes a recent trend to propose security-defending techniques against adversarial patches, including heuristic defenses and certified defenses. 
Heuristic defenses~\cite{naseer2019local,rao2020adversarial, chen2021turning, chou2020sentinet, rossolini2022real} are mostly based on the empirical understanding of existing attacks. 
{\responserefa{}
    Among them, empirical adversarial detection shares a similar setting with us. The problem is formulated as a binary classification problem. Some approaches utilize another machine learning model (such as DNN) to classify the inputs as natural or adversarial by directly analyzing the input~\cite{gong2017adversarial, metzen2017detecting, grosse2017statistical} or utilizing features extracted from the neural networks~\cite{carrara2018adversarial, wang2021smsnet, abusnaina2021adversarial}. Another line of work tries first to analyze the DNNs to understand their internals and leverage related knowledge for the detection~\cite{liu2018adversarial, du2019deepstellar, yang2020ml}. Finally, some papers focus more on the input and try to detect adversarial inputs through input reconstruction~\cite{meng2017magnet, samangouei2018defense} or feature squeezing~\cite{xu2017feature}. However, such empirical attacks may fail on strong adaptive attacks~\cite{chiang2020certified, carlini2017adversarial} and are thus unreliable in safety-critical scenarios. 
}

Certified defenses, instead, aim to rigorously guarantee the robustness of a given model to patch attacks. It is worth noticing that different from whole-image perturbations, where adversarial and natural perturbations are hard to analyze in a uniform framework, it is possible to unify both kinds of perturbations in the patch setting. In other words, given a fixed patch size, once the robustness against the worst-case adversarial case is guaranteed, the safety for natural perturbations can then be obtained also. Due to the promising capability to defend strong adaptive adversaries, certified defenses have attracted many researchers' interest in recent years.

\begin{figure}[tbp]
     \centering
     \includegraphics[width=0.85\linewidth]{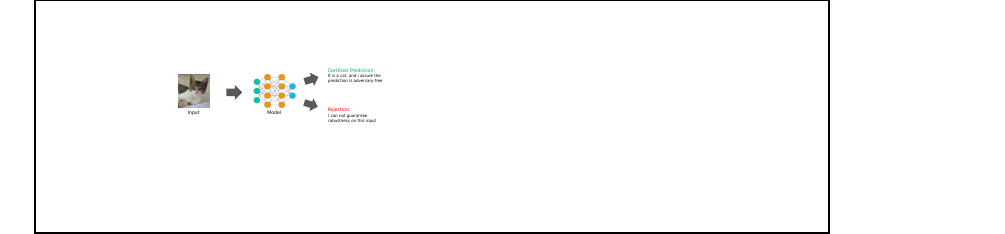}
     \caption{An illustration of detection-based certified defense.}
     \label{fig:bg:illustration}
\end{figure}

At a high level, existing certified defenses mostly fall into two categories: \emph{certified recovery} \cite{chiang2020certified,xiang2020patchguard,levine2020randomized,zhang2020clipped,metzen2021bagcert, salman2021certified} and \emph{certified detection} \cite{mccoyd2020minority, xiang2021patchguard++, han2021scalecert, xiang2022patchcleanser}. 
Certified recovery aims to recover the correct prediction on images with adversarial patches, while certified detection aims to detect adversaries with a provable detection rate, as shown in Figure~\ref{fig:bg:illustration}. 
{\responserefa{}In the following, we briefly introduce three representative certification workflows that are closely related to ours, two for certified recovery and one for certified detection. Notice that the success of the patch attack comes from the non-linear property of the neural networks: a small local area can influence the prediction of the global input. As a result, these works share a similarity that they all try to isolate or block the influence of the adversarial patch through different perspectives. 

(De)Randomized Smoothing (PatchSmoothing)~\cite{levine2020randomized} is one of the early attempts that belongs to the certified recovery category. Its main contribution is to select only part of the image as one input and perform multiple inferences across the whole image. There are two selection strategies in the paper which are called block smoothing and band smoothing. The block smoothing chooses a square with size $s\times s$ as one input and enumerates all $h \times w$ positions where $h$ is the height of the image, and $w$ is the width of the image. When the center of the block hits the boundary of the image, PatchSmoothing wraps around (\eg use the pixels in the other three corners if the center is positioned in one corner). After counting the inference results of all these squares, PatchSmoothing computes the difference between top-2 classes, and if:

\begin{equation}
    \Delta \geq 2(m+s-1)^2
\end{equation}

Then, PatchSmoothing certifies the input with the label of top-1 class. Here $\Delta$ means the difference and $m$ is the size of the adversarial patch, $s$ is the square size. This threshold is decided by the maximum number of patches the adversarial can control. Band smooth works similarly, but instead, choosing a column with size $s$ as the input and the threshold is:

\begin{equation}
    \label{eq:PS_column_threshold}
    \Delta \geq 2(m+s-1)
\end{equation}

PatchGuard~\cite{xiang2020patchguard} moves one step further. It utilizes a particular CNN  called bagnet~\cite{brendel2019bagnet}, which has small receptive fields. The restricted receptive fields limit the effects of the adversarial patch. Based on such a structure, PatchGuard proposes an element-wise linear aggregation method to replace the original unsecured fully-connected layer, which will mix both benign and adversarial patches' prediction results. The aggregation mechanism works as follows: It checks the prediction vectors of all the sliding windows with a predefined size (adversarial patch size). If a sliding window with the highest score with respect to one class exceeds a predefined threshold, PatchGuard will mask that sliding window and uses the remaining window to predict one class. Such a mechanism is motivated by the observation that the adversarial patch will increase the score of a target class to an extremely high value to surpass the original prediction. 

These two certified recovery methods both utilize the voter mechanism, while the former ensures that the adversary won't influence the majority, and the latter ensures the suspicious voter can't control the result. We detail these two methods that are parallel to our work because, in this paper, we want to discuss the difference between certified recovery and certified detection and provide both practical concerns and experiment results for our choice of certified detection. 

Minority Report (MR)~\cite{mccoyd2020minority}, on the other hand, is a representative work of certified detection. For every input, it also makes multiple inferences where each one is masked by an occlusion region $s \times s$ that is two pixels larger than a predefined adversarial patch. The occlusion region slides across the image (with size $h \times w$), which requires MR to collect results for $(h - s + 1) (w - s + 1)$ inference. MR only certifies the result if all predictions yield the same class. While in their implementation, this requirement is relaxed by first discarding the lowest score to tolerate outliers and providing certified results when the supported votes exceed a predefined threshold. Such relaxation will incur false negatives but increase the overall performance.

PatchCleanser~\cite{xiang2022patchcleanser} is one recent work as a certified detection. At the time we finished the draft of the paper, we were unaware of its existence. It could be considered an expanded version of our method beyond the transformer. However, our work differs from theirs in that we offer a more comprehensive and in-depth examination of the difference between certified detection and certified recovery, encompassing the small AOI and strong adversary scenarios. We also devote a significant amount of space to discussing the real-world implementation of certified detection.
}

The focus of this paper is certified detection. \sys differentiates existing certified detection methods in three ways:

\begin{itemize}[leftmargin=*]
    \item {\sys is built on an exhaustive testing strategy and does not require extra effort to train a model with patch robustness.}
    
    \item {Certification of \sys works on a range of patch sizes instead of one fixed size, which is more flexible.}
    
    \item {\sys can be applied for patch sizes up to 25\% (while most related work is evaluated on 2\%-pixel attack), which could be more practical when detecting natural patch perturbations. }
\end{itemize}

\section{Motivational Study}
\label{sec:motivation}

Certified recovery and certified detection for adversarial patches both aim to improve the robustness of DNNs and provide a formal guarantee. In this section, we present preliminary experiments to show the shortcomings of certified recovery and our motivation behind choosing certified detection. 

Most existing certification approaches for adversarial patch defense are focused on certified recovery.
Such approaches are typically based on a special model (\eg BagNet\cite{brendel2019bagnet}) that gives predictions with small receptive fields.
They mostly require splitting the input image into small regions, each of which is fed into the model to generate a local prediction.
The defended prediction is made by voting over all the local predictions, and the certification for an input sample is obtained if sufficient local predictions vote for the same label, ensuring that an arbitrary adversarial patch would not change the prediction. In other words, given $\cY$ classes, the image is certified as $class_A$ if the following condition is satisfied:

\begin{align}
    \begin{split}
        \label{eq:voting}
        Vote_{class_A} - Vote_{class_B} &> threshold \\
        \text{where} \quad Vote_{class_B} &= max_{\forall i \in \cY, i \neq A}(Vote_{class_i})  
    \end{split}
\end{align}

The threshold differs between different methods. 


Although certified recovery is usually expected to provide a more rigorous and formal guarantee, we believe that offering a lightweight and statistical guarantee for whether there is an attack may be more practical for real-world applications.

\begin{itemize}[leftmargin=*]
    \item First, one obvious shortcoming of certified recovery is its limited performance (e.g., only 26.0\% certified accuracy on ImageNet for a recent work~\cite{metzen2021bagcert}). This often makes such defense impracticable for real-world scenarios. 

    \item Second, optimizing the model for both clean and adversarial inputs would inevitably decrease the model's accuracy in all existing certified recovery approaches, which may harm the user experience in normal settings. However, it is hard for the current certified recovery method to achieve moderate certified performance without such optimization. Besides, such optimization, usually done by re-training or even training from scratch, may take more time and computing resources. 

    \item In addition, we observe that recovering the correct prediction under the presence of adversarial patches is difficult or even impossible in some cases, \eg when the critical {region of interest (ROI)} is hidden by an adversarial patch. Such a mechanism may be difficult to deal with images with a small {ROI}, where most local voters may not get enough information to produce the correct prediction.
    With the advance of camera technology, it is more common to have high-resolution pictures nowadays, leading to a much smaller {\responserefa{}ROI} in practice than the popular MNIST and CIFAR-10 datasets.

\end{itemize}

To further validate our observation and hypothesis, we performed a motivational study by conducting studies to investigate two state-of-the-art certified recovery approaches, i.e., PatchGuard~\cite{xiang2020patchguard} (PG for short) and De-randomized Smoothing~\cite{levine2020randomized} (DS for short).

\begin{figure}[tbp]
     \centering
     \includegraphics[width=0.5\linewidth]{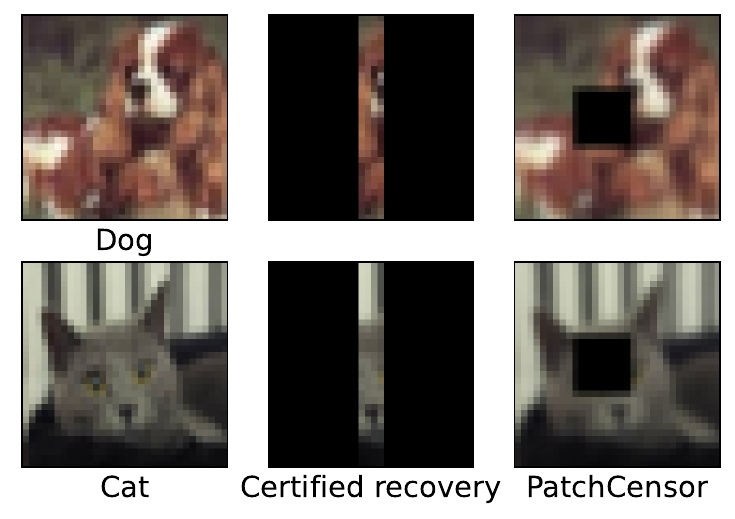}
     \caption{An illustration of the information used by different defense approaches for prediction and certification. Certified recovery approaches are based on small local regions. Our method is based on the whole image with a small region occluded.}
     \label{fig:motivation:pieces}
 \end{figure}
 
\begin{figure}[tbp]
     \centering
     \includegraphics[width=0.4\linewidth]{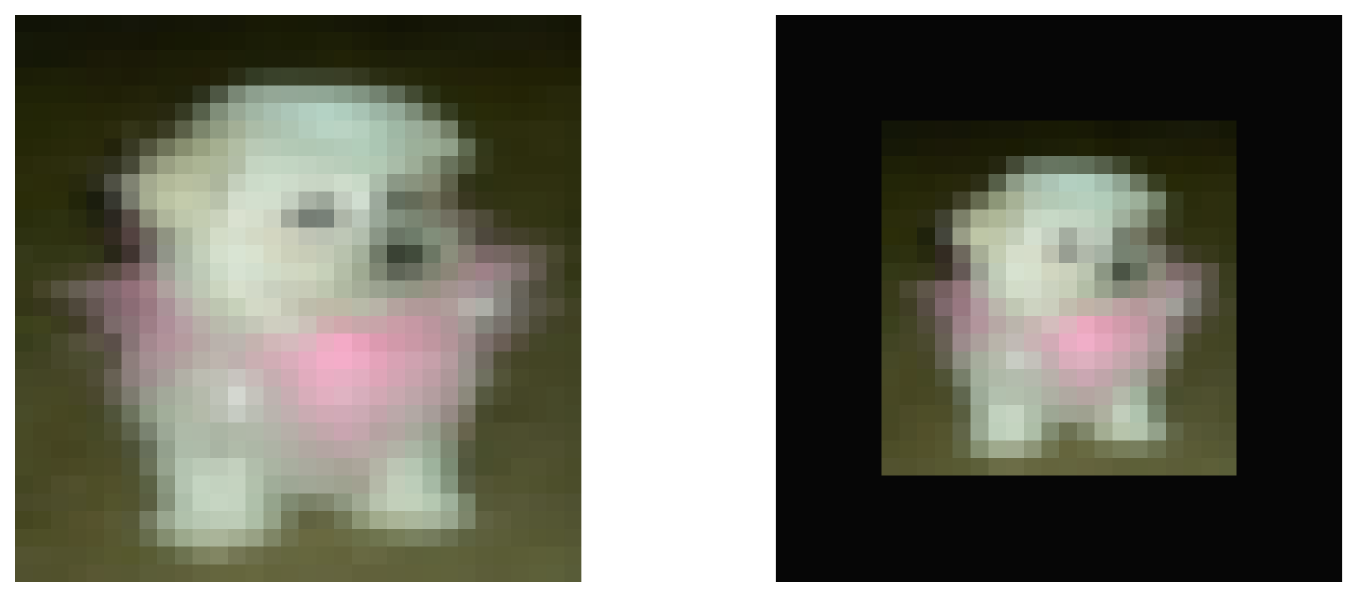}
     \caption{An illustration of image rescaling. The left one is the original image, and the right one is the image after rescaling.} 
     \label{fig:motivation:resize}
 \end{figure}

\begin{figure}[]
\centering
\begin{minipage}{0.45\linewidth}
    \centering
    \includegraphics[width=2.4in]{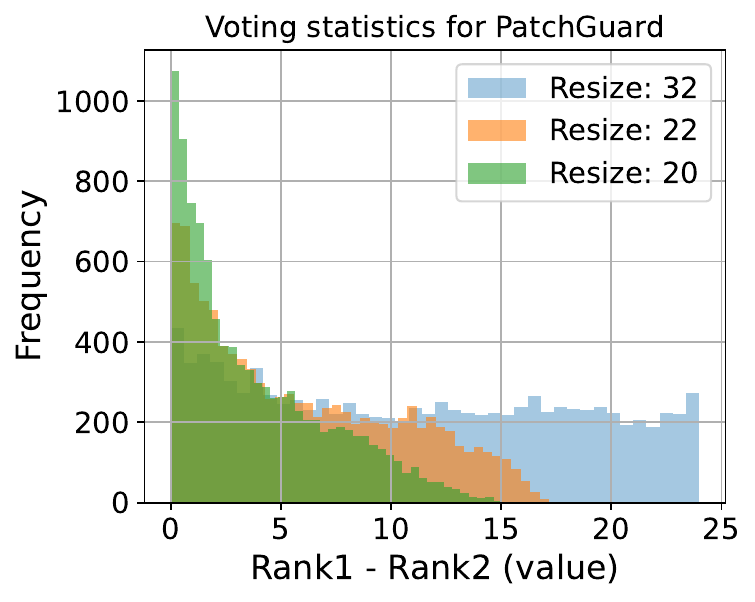}
\end{minipage}
\hspace{0.2cm}
\begin{minipage}{0.45\linewidth}
    \centering
    \includegraphics[width=2.4in]{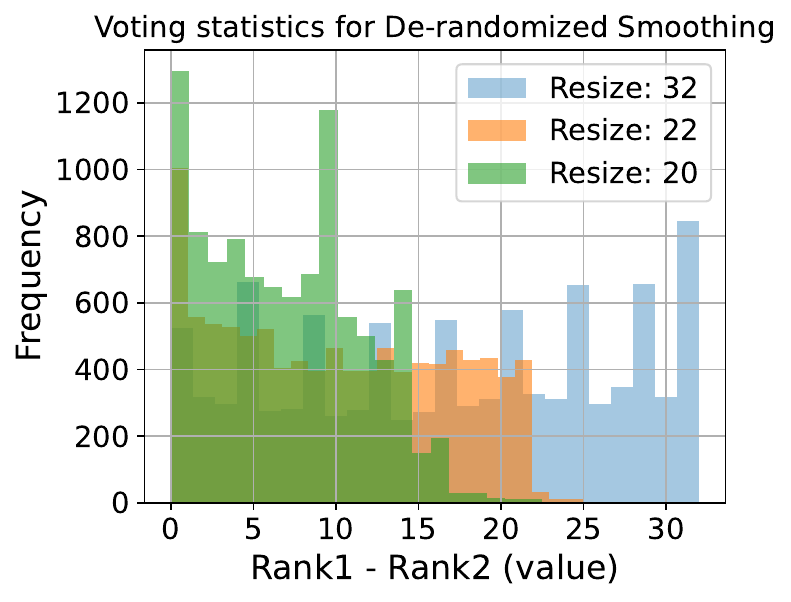}
\end{minipage}
\caption{The voting distribution of PatchGuard and De-randomized Smoothing with different rescaling sizes. X-axis shows the value difference between top-2 voting. Y-axis is the number of samples that produce the voting result. 
}
\label{fig:motivation:voting_distribution}
\end{figure}

Specifically, we rescale each image in CIFAR-10 to a smaller size and pad the image to the original size, as shown in Fig.~\ref{fig:motivation:resize}. In this way, we can control the {\responserefa{}ROI} size of each image. Since the images after resizing are different from the original ones, we have to retrain the models. We train the De-randomized Smoothing ResNet (DS-ResNet) from scratch for 200 epochs, using the common settings in De-randomized Smoothing~\cite{levine2020randomized} and PatchGuard~\cite{xiang2020patchguard}.
We evaluate these two methods with 2.4\% adversarial patch size and record the distribution shift of the voting differences ($Vote_{Rank1\_class} - Vote_{Rank2\_class}$), as shown in Figure~\ref{fig:motivation:voting_distribution}.

It is obvious that when the ROI becomes smaller, the margin between Rank-1 voting and Rank-2 voting becomes smaller. If the voting margin of a sample is below a threshold, the model does not yield a certified result for the sample.
The threshold can be understood more intuitively for randomized Smoothing (as PatchGuard is based on the weighted sum and the voting value is continuous), in which the threshold is computed by the equation~\ref{eq:PS_column_threshold}.

For example, if the $m$ (target size to certify) is 5 (2.4\% patch size), and the $s$ (column size) is 4 (default setting), then the threshold is 16. In Figure~\ref{fig:motivation:voting_distribution}, we can clearly see that the distribution of De-randomized Smoothing for size 20 has just shifted to the left of 16. As a result, there is a sharp drop in the certified accuracy at size 20. The detailed results can be found in Figure~\ref{fig:evaluation:resize_accuracy} (see section ~\ref{sec:results}).

\begin{figure}[tbp]
     \centering
     \includegraphics[width=0.85\linewidth]{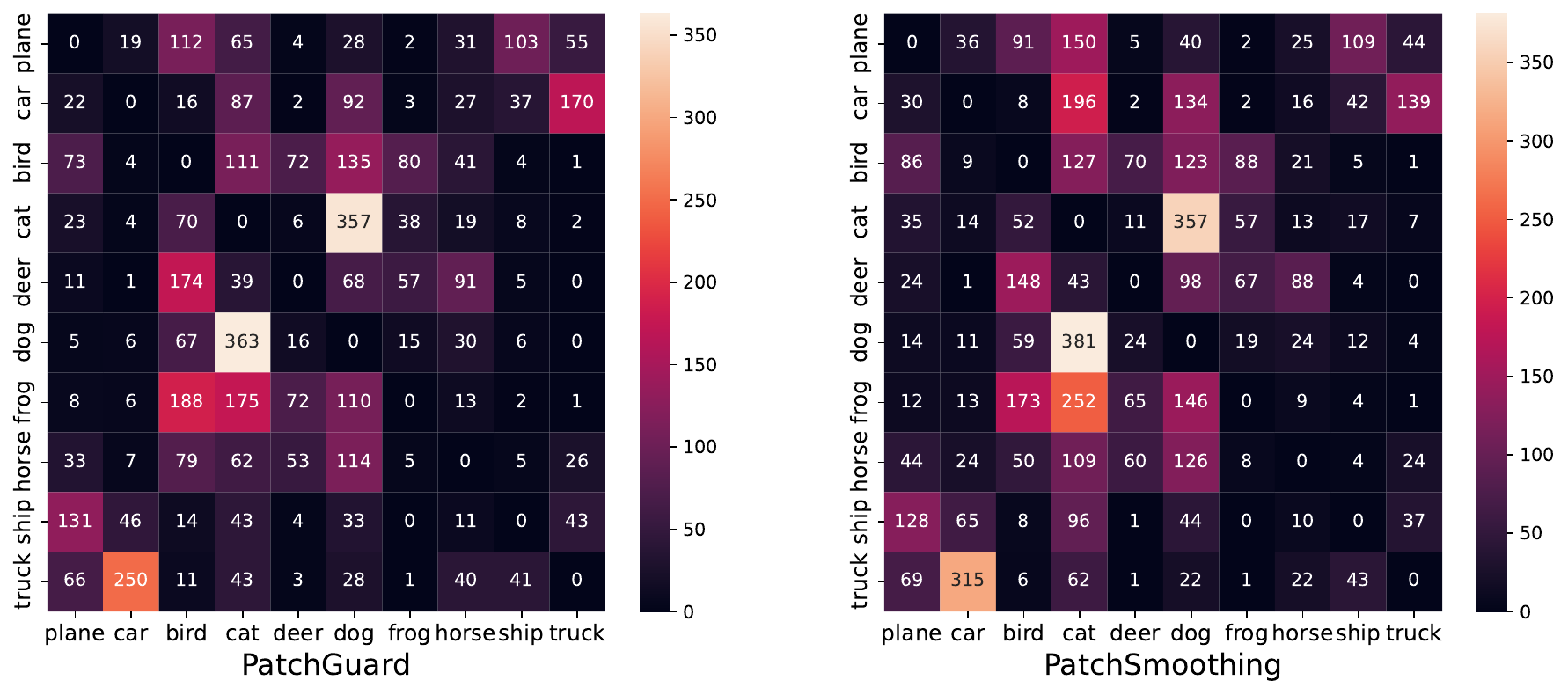}
     \caption{Confusion matrix for test instances that are unable to certify but predicted correctly. Y-axis indicates the ranked-1 class and X-axis indicates the ranked-2 class}
     \label{fig:motivation:confusion_matrix}
 \end{figure}

We additionally analyze test instances that are predicted correctly but unable to certify in the CIFAR-10 dataset to show that taking only local features as input may cause confusion for certified recovery methods. In other words, the difference between the top-2 classes of these instances does not exceed the threshold defined in these algorithms. The confusion matrix for those instances is shown in Fig.~\ref{fig:motivation:confusion_matrix}, where the y-axis indicates the true class, and the x-axis indicates the ranked-2 class. We can clearly find that for both methods, dog \vs cat, car \vs truck and frog \vs bird are classes that are hard to certify. And these classes are sometimes hard to distinguish even for humans, given only a small patch of the image, as shown in Fig.~\ref{fig:motivation:pieces}

The results of our preliminary study confirm that the certified recovery methods relying on local features could have fundamental limitations in dealing with small-ROI images, making them difficult to achieve reasonable certified accuracy in practice.

On the other hand, using as many patches as possible in the inference pass can provide sufficient information for the models to predict. This can improve clean accuracy, especially under a small ROI. However, increasing the number of patches in one inference will inevitably increase the chance of including malicious patches and, as a result, increase the difficulty for the defense. Our work aims to alleviate this dilemma by proposing a defense mechanism in a detection manner which can use more patches to improve certified accuracy.







\section{\sys: CERTIFIED DETECTION VIA EXHAUSTIVE TESTING}
\label{sec:method}

In this section, we originally proposed a testing-based certified defense technique against adversarial patches, named \sys. We first formulate the problem and formally define the properties that we intend to certify; then we present our technique and make high-level theoretical analysis.

\subsection{Problem Formulation}

To be general, our work shares the same threat model as existing recent state-of-the-art certified defense work against adversarial patches \cite{xiang2020patchguard,levine2020randomized,chiang2020certified,zhang2020clipped}
(The attack and defense are both focused on image classification context.).
We use $\cX \subset \reals^{W \times H \times C}$ to denote the distribution of images where each image $x \in \cX$ has width $W$, height $H$, number of channels $C$. We take $\cY=\{0,1,\cdots,N-1\}$ as the label space, where the number of classes is $N$. 
We use $f: \cX \rightarrow \cY$ to denote the model that takes an image $x \in \cX$ as input and predicts the class label $y \in \cY$.


\textbf{Attacker capability.} The attacker can arbitrarily modify pixels within a restricted region, and this region can be anywhere on the image, even over the salient object. We assume that all manipulated pixels are within a square region, and the defender has a conservative estimate (\ie, upper bound) of the region size. Our technique can be generalized to other patch shapes, as long as the patch can be covered by a restricted rectangle, but we focus on square-shaped patches for simplicity.

Formally, we assume the attacker can arbitrarily modify an image $x$ within a constraint set $\cA(x)$. We use a binary pixel block $p \in \cP \subset \{0,1\}^{W \times H}$ to represent the restricted region, where the pixels within the region are set to $1$.
Then, the constraint set $\cA(x)$ can be expressed as $\{\xp = (1-p) \odot x+p \odot \xpp\} | x, \xp \in \cX, \xpp \in \reals^{W \times H \times C}, p \in \cP\}$, where $\odot$ refers to the element-wise product operator, and $\xpp$ is the content of the adversarial patch.

\textbf{Attack objective.} We focus on adversarial patch attacks against image classification models. Given a deep learning model $f$, an image $x$, and its true class label $y$, the goal of the attacker is to find an image $\xp \in \cA(x) \subset \cX$ such that $f(\xp) = y'$, where $y'$ is an arbitrary incorrect class label defined by the attacker and $y' \neq y$.


\textbf{Defense objective.} The role of the defender is to design a defended model $g = (f,v): \cX \rightarrow \cY \times \{0,1\}$, where $g(x) = f(x),v(x)$, $f(x) \in \cY$ is the classification result, and $v(x) \in \{0,1\}$ is the verification result indicating whether the prediction $f(x)$ can be verified ($1$ stands for ``verified''). A verified inference means this inference is not influenced by any attacker, and we can trust the prediction made by the model.

\textbf{The property to be certified.} Based on the defense objective, we formulate the property to be certified as follows: Given a model $M$ and a data distribution $D$, we certify that for any $x \in D$, the prediction $M(x)$ is correct and robust (cannot be falsified by an arbitrary attacker with ability $A$) with a probability of $\theta$.

Certification for neural networks in the form of statistical guarantee is not a new idea. Most certified defending techniques in the digital attack domain~\cite{lecuyer2019certified, cohen2019certified} also offer such a guarantee. However, this formulation may cause confusion since most previous certification work (including certified detection, which is the topic and focus of this paper) on patch attacks offers determined certification results. On the one hand, we find that for \sys and all the previous certified detection work is, in reality, offering such statistical guarantee (discussed in Section~\ref{sec:discussion}), and we believe it is necessary and important to reformulate the problem. On the other hand, we argue that such a statistical guarantee is more practical in real-world applications. We will discuss more on this later.




\textbf{Evaluation metric.}
The defender aims to improve the quality of the defended model by improving clean accuracy and certified accuracy. 
The clean accuracy $\accClean$ (the accuracy of $f$ on the original dataset $\cX$ without considering verification\footnote{Please notice our clean accuracy definition differs from previous certified detection work. We report the $\accClean$ so the readers can understand the influence of pre-training for the base model.}) and the certified accuracy $\accCertified$ (the ratio of images in $\cX$ that are correctly and provably classified) are frequently used in prior work to describe the performance of certified defenses. Formally, 
\begin{equation}
    \accClean = \bE_{x \in \cX} [l(f(x), y)]
\end{equation}
\begin{equation}
    \accCertified = \bE_{x \in \cX} [l(f(x), y)\ v(x)]
\end{equation}
where $l$ is the 0/1 error function.

In other words, the defender intends to maximize the ratio of verifiable inputs, measured by
\begin{equation}
    \rTrust = \frac{|\cX_{trust}|}{|\cX|}
\end{equation}
and the model accuracy in the trust domain
\begin{equation}
\accInTrust = \bE_{x \in \cX_{trust}} [l(f(x), y)]
\end{equation}
so that it can retain high test accuracy and guarantee reliability on most inputs ($\cXtrust$), while raising a warning if the input $x$ is potentially malicious ($v(x) = 0$).
The concept is similar to the selective prediction approaches that try to integrate a reject option in the neural network \cite{geifman2019selectivenet}.
Such a rejection mechanism is helpful in many perception tasks in general where a fall-back solution is available. For example, a driving assistance model can ask the human driver to take over or perform a conservative operation when it's not confident about the current situation.

\subsection{The Testing-based Defense}


\begin{figure*}[]
    \centering
    \includegraphics[width=0.99\linewidth]{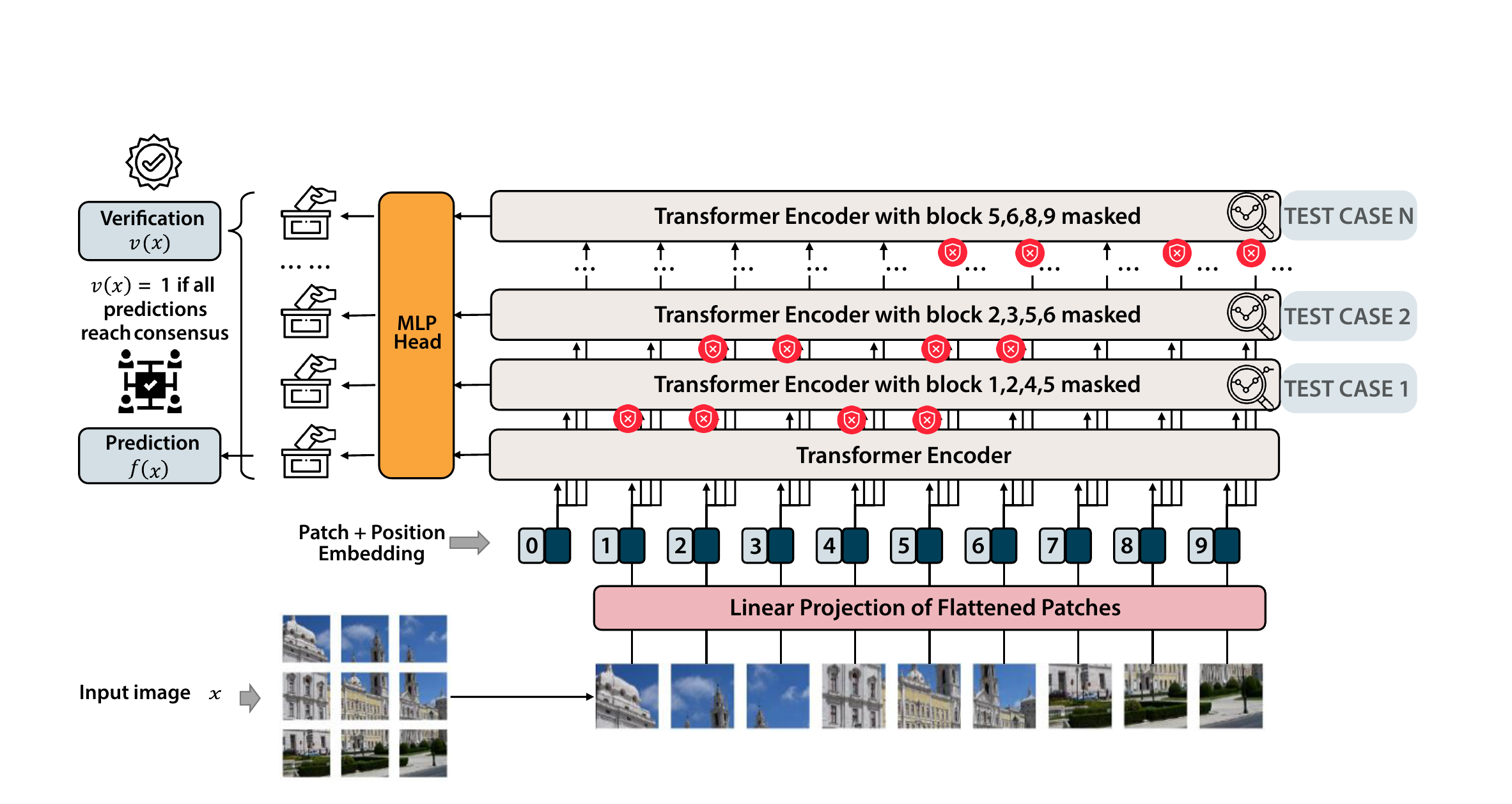}
    \caption{The overall architecture and workflow of \sys.}
    \label{fig:approach:arch}
\end{figure*}

Our adversarial patch defense technique is designed upon the Vision Transformer (ViT) architecture, by utilizing its input partition nature and self-attention mechanism.

The summarized workflow of \sys are shown in Figure~\ref{fig:approach:arch}.
It can be viewed as a paired function $g = (f,v)$, where $f$ is a pretrained ViT model for prediction, and $v$ is a verification function based on the ViT model.
Given an input image $x$, \sys produces a pair $(f(x), v(x))$, where $f(x)$ is the predicted class of $x$ and $v(x)$ represents whether the prediction is verified.

In \sys, the input image is split into non-overlapping patches as in vanilla ViT models.
Suppose the input patch size is $P \times P$ in pixels, then the input image $x \in \cX \subset \reals^{W \times H \times C}$ is partitioned into a sequence of patches $\cP = \{p_i | i=1, 2, ..., n\}$, where $n = n_w \times n_h = \frac{W}{P} \times \frac{H}{P}$ is the number of patches.
Each patch $p_i$ is then flattened, passed through a linear projection layer, and added a position embedding to generate a patch embedding $q_i$. A learnable \texttt{[class]} embedding $q_0$ is prepended to the sequence of embeddings, which is used to produce the class prediction after passing through later layers.

Specifically, the patch embeddings are fed into multiple parallel Transformer encoder layers $\cT = \{t_0, t_1, ..., t_k\}$ to exchange information between local patches. The encoders share the same weights of the ViT model $f$, while being used with different attention masks.
Each Transformer encoder $t_j$ produces an encoding of the \texttt{[class]} node, which is then passed through the MLP head to produce a class prediction $y_j$. We call the predictions produced with the masked attention maps (\ie $y_1, y_2, ..., y_k$) as \emph{masked predictions}.
To test whether there is an adversarial patch, we design a special mutation operator that directly masks the attention maps. The predictions with different masks can be seen as different mutations. The presence of the patch can be detected once we find anomalies among the test results. We call the predictions produced with the masked attention maps (\ie $y_1, y_2, ..., y_k$) as \emph{mutations}. 
The class prediction $f(x)$ is produced by the Transformer encoder without mutation (\ie $y_0$), which is equivalent to a direct inference using the base ViT model.
The verification result $v(x)$ is produced by voting over all of the testing results ($y_1$ to $y_k$). We ensure that at least one of the mutations is benign (\ie can completely mask the adversarial patch out), which vetoes the adversary's target output even if all other predictions are compromised. The prediction is verified if all Transformer encoders reach consensus (\ie, agreeing on the same class prediction).
As we enumerate all possible positions for the potential adversarial patch in this test generation process before masking, this test can exhaustively cover all corner cases, meaning that we complete a \emph{full-coverage testing}. 

Ideally, if testing can be done on all possible situations and these test cases are all passed (producing correct and consistent output), the exhaustive testing would be equivalent to a rigorous verification. However, directly testing all possible adversarial patch positions is computationally infeasible and impractical considering its combinatorial complexity.
Our mask strategy plays a role similar to the abstract set in abstract interpretation~\cite{cousot1977abstract}, providing a convenient (due to the natural robustness of self-attention architecture) and efficient (due to the reduced search space) way to achieve exhaustive full-coverage testing.  

As a concrete example, we illustrate our method in Figure~\ref{fig:approach:arch}. Suppose we use a ViT model with 30$\times$30 input resolution and 10$\times$10 input patch resolution as the base model.
An input image will be partitioned into 3$\times$3 patches. To defend against adversarial patches with 5$\times$5 resolution, we let each mask exclude 2$\times$2 local patches, \ie, a square region with 20$\times$20 resolution. By sliding the 2$\times$2 mask over all local patches in the image, we can obtain four (2$\times$2) possible mask locations and guarantee at least one of the mask locations can completely hide the 5$\times$5-pixel adversarial patch.
A certified prediction is produced if the four masked predictions vote for the same class as the non-masked prediction.



\subsection{Mutation Strategy}
\label{sec:masking}

\begin{figure}[tbp]
    \centering
    \begin{subfigure}[b]{0.33\linewidth}
        \centering
        \includegraphics[width=1.0in]{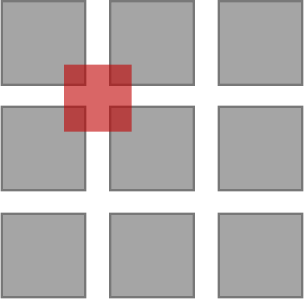}
        \caption{$0 < W_{adv} < P$}
    \end{subfigure}
    ~
    \begin{subfigure}[b]{0.33\linewidth}
        \centering
        \includegraphics[width=1.0in]{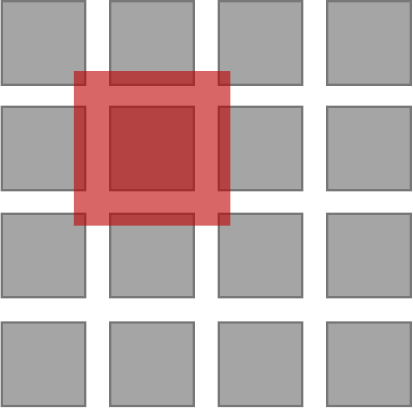}
        \caption{$P < W_{adv} < 2P$}
    \end{subfigure}
    ~
    \begin{subfigure}[b]{0.33\linewidth}
        \centering
        \includegraphics[width=1.0in]{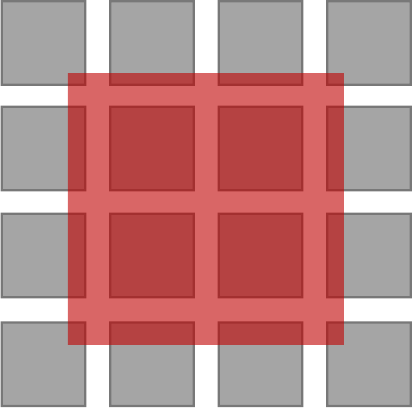}
        \caption{$2P < W_{adv} < 3P$}
    \end{subfigure}

    \caption{Number of ViT patches affected by different sizes of adversarial patches. $P$ and $W_{adv}$ represent the widths of the ViT input patch and adversarial patch respectively.}
    \label{fig:approach:patch_examples}
    \vspace{-5mm}
\end{figure}

The certification capability of \sys is achieved by testing with different mutations, while ensuring that at least one of the mutations can lead to an uninfluenced prediction (\ie, the corresponding attention mask can exclude the adversarial patch). The idea is analogous to Byzantine Fault Tolerance~\cite{castro1999practical} where the benign voter can control the final result.


Each mutation leverages a different attention mask for inference. The key to designing the attention masks is to determine how many local input patches might be tainted by the adversarial patch (\ie containing pixels belonging to the adversarial patch).
Since the input partition plan is fixed and the adversarial patch is at an arbitrary location in the image, we need to consider the worst case, \ie the maximum number of input patches that may be tainted by the adversarial patch.

Our masking strategy is based on the observation that, 
\emph{in an image that is partitioned into non-overlapping $P \times P$-pixel patches, an adversarial patch with $W_{adv} \times H_{adv}$ resolution can affect at most $N_{W} \times N_{H}$ input patches, where $N_{W} = \lceil \frac{W_{adv}}{P}\rceil + 1$ and $N_{H} = \lceil \frac{H_{adv}}{P}\rceil + 1$.}
For example, as illustrated in Figure~\ref{fig:approach:patch_examples}, if the width $W_{adv}$ of a square adversarial patch is smaller than the input patch width $P$, four input patches may be tainted (if the adversarial patch is at the joint of 2$\times$2 input patches). Similarly, a square adversarial patch with $W_{adv} \in (2P, 3P)$ may affect at most 4$\times$4 input patches.

{\responserefa{}
We would like to emphasize that our masking strategy here ensures that for any object at any location with any shape, as long as it can be restricted to a rectangle with shape $N_W \times N_H$, there must be a masking that can mask this object out. This flexibility makes it possible to detect both adversarial patches and abnormal objects in practice, while previous work often requires fine-tuning the model for a specific patch size.
}

Specifically, given the maximum rectangle shape $W_{adv} \times H_{adv}$ of the adversarial patch, we can compute the minimum size $N_{W} \times N_{H}$ for the attention mask. By sliding the mask over the whole input patch grid with a stride of 1, we can enumerate all $k$ possible locations of the mask, where
\begin{equation}
    \label{equation:num_masks}
    k = (\frac{W}{P} - N_{W} + 1) \times (\frac{H}{P} - N_{H} + 1)
\end{equation}
We can guarantee that at least one of the $k$ masks can cover the arbitrary adversarial patch.

 
As we only mask a small proportion of the image and our base model is the powerful ViT, it is relatively easy for all voters to reach a consensus for a correct prediction on clean data.



\subsection{Certification Analysis}
\label{sec:certification}

In this subsection, we provide analysis to show that \sys can achieve the defender's objective described in Section~\ref{sec:motivation}.

After obtaining the regular prediction of the ViT base model $f(x)$, the verification result of \sys is obtained by
\begin{eqnarray}
    \label{eq:verification}
    v(x) \triangleq \left\{
                 \begin{array}{ll}
                 1, & \hbox{if $f_1(x)=f_2(x)=...=f_k(x)=f(x)$;} \\
                 0, & \hbox{otherwise} 
                 \end{array}
               \right.
\end{eqnarray}
where $f_j(x)$ represents the prediction obtained by the ViT base model with the $j$-th mask position on the attention map.

For any verified clean data point $x \in \cX_{trust}$ and any adversarial example $\xp \in \cA(x)$, we need to ensure the adversarial patch is either ineffective or can be detected.
Specifically, assuming $\xp$ can pass the verification, we have
\begin{equation}
f_1(\xp)=f_2(\xp)=...=f_k(\xp)=f(\xp)   
\end{equation}
Based on our masking strategy (Section~\ref{sec:masking}), at least one of the masked predictions will exclude the adversarial patch in $\xp$, \ie
\begin{equation}
\exists j \in \{1, 2, ..., k\},\ \textit{s.t.}\ f_j(\xp) = f_j(x)   
\end{equation}
Since $x$ is a verified input, we have $f(x) = f_j(x)$, and thus
\begin{equation}
f(\xp) = f_j(\xp) = f_j(x) = f(x) 
\end{equation}
meaning the adversarial patch attack on $\xp$ is ineffective. 

However, it is still possible that the benign voter(s) make mistakes. This will eventually lead to an error as described in Section~\ref{sec:discussion}. However, such an error originates from the insufficient capability of the base model and is not influenced by the attacker. {\responserefa{ This error can also be estimated before model implementation as long as the data distribution for the benign patches (\ie patches that do not contain OOD objects or are not adversarially influenced) is similar before and after implementation.} Notice that the similar distribution assumption is one of the key assumptions in statistical machine learning theory~\cite{vapnik1999nature, kawaguchi2017generalization}} 

The estimated upper bound of the proportion of instances that the benign voters make mistakes is just $(1-\accCertified)$ on clean data. In other words, the probability of a correct and robust inference $\theta$ as we mentioned earlier is $\accCertified$. This estimated confidence can serve as an indicator of how much we can trust the model when the full-coverage testing is passed.




\section{Evaluation}
\label{sec:results}

{\responserefa{}
To better understand the performance of \sys and its relation with state-of-the-art (SOTA) techniques from different angles, we design four research questions. 

We first investigate the performance of our method and compare it with both certified recovery and certified detection methods in RQ1:

{\noindent \textbf{RQ1: How does the proposed system \sys perform in terms of clean accuracy and certified accuracy?}}

This research question aims to offer an overview picture for the comparison of all the methods we have mentioned so far. To answer this research question, we select CIFAR-10~\cite{krizhevsky2009learning} and ImageNet~\cite{krizhevsky2012ImageNet} datasets under different adversarial patch sizes as our subject datasets for evaluation, which are also widely used in previous studies on adversarial patch defense. In addition to these two default evaluation datasets in related literature, we also conduct experiments on the German Traffic Sign Recognition Benchmark (GTSRB) dataset~\cite{stallkamp2011german} and Food-101 dataset~\cite{bossard2014food}}. Resolution and number of instances usually reflect the difficulty of the dataset. We choose these four datasets to represent different levels of CV tasks. Specifically, CIFAR-10 contains 60,000 images with $32\times32$ resolution, GTSRB contains 51,839 images with $48\times48$ resolution, Food-101 contains 101,000 images with $384\times384$ resolution, and ImageNet contains 1,431,167 images with average resolution $469\times387$. We report the clean accuracy and certified accuracy with 2.4\% adversarial patch size on CIFAR-10 and 2\% on ImageNet GTSRB and Food-101 dataset, following the setting in the previous work~\cite{chiang2020certified, levine2020randomized, mccoyd2020minority, metzen2021bagcert}. 

After the general evaluation, we would like to verify the observations from our motivation study and discuss the difference between our methods (certified detection) and certified recovery when the area of ROI is small in RQ2:

{\noindent \textbf{RQ2: Does the proposed system \sys achieve moderate certified performance even for images with a small area of the region of interest (ROI)?}}

Specifically, the aim of RQ2 is to discuss the reason behind choosing certified detection instead of certified recovery (as mentioned in the motivation study). Discussion on this decision is an important focus of this paper, and RQ2 serves as a foundation for our further analysis in section 6. 
It is also worth noting that we have compared the performance of \sys with MR+ in RQ1 and showed that \sys achieves SOTA as a certified detection method. In RQ2, we aim to compare certified detection with certified recovery at the category level. Therefore, we select the most competitive ones from each category and did not include the second-best MR+ in this RQ.

After the discussion on certified detection and certified recovery, we then would like to offer an in-depth analysis of certified detection. Our evaluation is thus turned to the comparison between \sys and the previous SOTA, MR+. One advantage of certified detection is its ability to defend the adversarially controlled patch with extremely large sizes (\eg up to 25\% of the whole image). So in RQ3, we want to investigate the performance of \sys under such strong adversaries on CIFAR-10, GTSRB, Food-101 and ImageNet: 

{\noindent \textbf{RQ3: How does the proposed \sys perform under strong adversaries?}}

Specifically, we increase the adversarial patch size from 0.5\% to up 25\%. Notice that it is possible to detect natural abnormal objects under such a setting. Also, when the size of dropped patches is large, fine-tuning may greatly influence the models' performance. So we also discuss the influence of fine-tuning in this RQ. We do not report the results of certified recovery because they are incapable of dealing with such large adversarial patches. In our experiments, the training is even hard to converge when the patch size is large. 

Finally, performance improvement usually comes at a cost. We intend to understand what is the computation overhead for \sys across different adversarial patch sizes compared with previous state-of-the-art MR+. The overhead can be divided into two parts: the training time and the inference latency. In this RQ, we mainly discuss the inference latency as it is the most important factor in practical implementation:

{\noindent \textbf{RQ4: What is the computational overhead of \sys compared with the state-of-the-art certified detection technique?}}

{\responserefa{}Notice that according to the mechanism we described in section~\ref{sec:method}, the larger the target patch size, the fewer inference needed for \sys. There is some trade-off in this relation as a large patch will yield worse certified accuracy but lower inference latency. To investigate such a relation, RQ4 includes experiments to guide choosing a suitable target patch size at runtime. We also provide experiments to demonstrate the influence of different ViT variants. Similar to RQ3, certified recovery methods will only provide trivial results when the patch size is large. Also, the fundamental certification goal is different, so we do not perform the evaluation in this RQ.}


\sys is implemented in Python based on a pre-trained ViT-Base model variant with $16 \times 16$ input patch size (ViT-B/16) \cite{rw2019timm}, which achieves 81.8\% test accuracy on ImageNet.\footnote{\scriptsize All the source code, documents and scripts are made available at the project website~\cite{patchcensor}.}

{\responserefa{}
    To support our large-scale evaluation, experiments were conducted on a Linux server with 4 Nvidia V100 GPUs. All the experiments take around 1,200 GPU hours. In the rest of this section, we summarize the key results for each of the studied research questions. Our code is released on the project website~\cite{patchcensor}.
}

\subsection{RQ1: Performance under Normal Setting}

\textbf{Experimental Settings.} To evaluate the proposed \sys under well-established settings, we compare \sys (PC for short) with eight state-of-the-art certified adversarial patch defense techniques on CIFAR-10 and ImageNet as these two are, by default, the commonly-used datasets in related literature. The compared methods include four recovery-based methods and four detection-based methods. Recovery-based methods include Interval Bound Propagation (IBP) \cite{chiang2020certified}, De-randomized Smoothing (DS) \cite{levine2020randomized}, PatchGuard (PG) \cite{xiang2020patchguard}, and BagCert (BC) \cite{metzen2021bagcert},
and detection-based methods include Minority Report (MR) \cite{mccoyd2020minority}, PatchGuard++ (PG++) \cite{xiang2021patchguard++} and ScaleCert (SC) \cite{han2021scalecert}.  
Among them, MR is based on directly training (fine-tuning) a CNN model to classify the images with a square region occluded, thus can achieve higher clean and certified accuracy.
However, the original design of MR needs to enumerate a lot of occlusion positions, which is computationally intensive for high-resolution images. Thus we additionally implement a more advanced version of MR, named MR+, by porting our masking strategy to reduce its number of occlusion positions to the same as in Equation~\ref{equation:num_masks}. 
We also replaced the CNN backbone of MR+ with a pre-trained ResNet50 \cite{rw2019timm} that has similar performance (81.1\% accuracy on ImageNet) as the ViT backbone ({\responserefb{}ViT-B/16}) used in \sys. 

{\responserefa{}
    After the experiments on CIFAR-10 and ImageNet, we further provide evaluations on GTSRB and Food-101. We include four methods: De-randomized Smoothing (DS), PatchGuard (PG), advanced Minority Report (MR+), and \sys. We choose these four methods because (1) they achieve competitive performance on CIFAR-10 and ImageNet; (2) they are representative methods in related areas; (3) they are open-sourced. 
}

\begin{table*}[]
    \centering
    \small
    \captionsetup{width=.9\linewidth}
    
    \caption{The clean and certified accuracy of different certified defenses on ImageNet and CIFAR-10. The numbers of IBP, DS, PG, BC, MR, PG++, and SC are directly copied from their paper. Note that the results of IBP and MR on ImageNet are not available, because they are computationally intensive or even infeasible on high-resolution images.}

    \begin{tabular}{cc|cc|cc}
    \toprule
    \multicolumn{2}{c|}{\multirow{3}{*}{Method}} & \multicolumn{2}{c|}{ImageNet} & \multicolumn{2}{c}{CIFAR-10} \\
    & & \multicolumn{2}{c|}{(2\% patch size)} & \multicolumn{2}{c}{(2.4\% patch size)} \\
                                  &                         & $\accClean$               & $\accCertified$              &  $\accClean$              & $\accCertified$               \\
    \midrule
    \multirow{4}{*}{Recovery}     & Interval Bound Propogation (IBP) \cite{chiang2020certified}    & \multicolumn{2}{c|}{N/A}                       & 47.8                 & 30.8                    \\
                                  & De-randomized Smoothing (DS) \cite{levine2020randomized}                     & 44.4                 & 14.0                    & 83.9                 & 56.2                    \\
                                  & PatchGuard (PG) \cite{xiang2020patchguard}             & 43.6                 & 15.7                   & 84.6                 & 57.7                    \\
                                  & BagCert (BC) \cite{metzen2021bagcert}               & 45.2                 & 22.9                   & 86.0                  & 60.0                     \\
    \midrule
    \multirow{9}{*}{Detection}    & \multirow{3}{*}{Minority Reports (MR) \cite{mccoyd2020minority}}                      & \multicolumn{2}{c|}{\multirow{3}{*}{N/A}}                      &92.5                 & 77.6                    \\
                                  & &                       &                      & 92.5                 & 62.1                    \\
                                  & &                       &                      & 92.5                & 43.8                    \\ \cline{2-6}
                                  & \multirow{4}{*}{PatchGuard++ (PG++) \cite{xiang2021patchguard++}}                      & 62.96 & 28.0 & 91.32                 & 68.9                    \\
                                  & & 62.96 & 32.0 & 91.32                 & 71.7                    \\
                                  & & 62.96 & 35.5 & 91.32                 & 74.3                    \\
                                  & & 62.96 & 39.0 & 91.32                 & 76.3                    \\
                                  \cline{2-6}
                                  & ScaleCert (SC) \cite{han2021scalecert}                     & N/A.\footnotemark[3]                & 55.4                   & N/A.\footnotemark[3]                 & 75.3                    \\
                                  & Minority Reports Adapted (MR+)                     & 75.5                 & 56.3                   & 97.7                 & 83.3                    \\
                                  & PatchCensor (PC, our approach)                   & \textbf{81.8}                 & \textbf{69.4}                   & \textbf{98.7}                 & \textbf{84.1}                   \\
    \bottomrule
    \end{tabular}
    ~
    \label{table:acc:baselines}
\end{table*}

\begin{table*}[]
    \centering
    \small
    \responserefa{}
    \captionsetup{width=.9\linewidth}
    
    \caption{{\responserefa{}The clean and certified accuracy of different certified defenses on GTSRB and Food-101. We fine-tune the models on GTSRB for 30 epochs and Food-101 for 60 epochs. All the models are fine-tuned with respect to the given patch size.}}

    \begin{tabular}{cc|cc|cc}
    \toprule
    \multicolumn{2}{c|}{\multirow{3}{*}{Method}} & \multicolumn{2}{c|}{GTSRB} & \multicolumn{2}{c}{Food-101} \\
    & & \multicolumn{2}{c|}{(2\% patch size)} & \multicolumn{2}{c}{(2\% patch size)} \\
                                  &                         & $\accClean$               & $\accCertified$              &  $\accClean$              & $\accCertified$               \\
    \midrule
    \multirow{2}{*}{Recovery}     
                                  & De-randomized Smoothing (DS) \cite{levine2020randomized}                     & 52.73                 & 15.97                    & 50.15                 & 17.33                    \\
                                  & PatchGuard (PG) \cite{xiang2020patchguard}             & 68.64                 & 33.61                  & 76.15                & 46.57                    \\
    \midrule
    \multirow{2}{*}{Detection}    
                                  & Minority Reports Adapted (MR+)                     & 96.36                 & 54.65                   & \textbf{84.68}                 & 64.80                    \\
                                  & PatchCensor (PC, our approach)                   & \textbf{99.89}                 & \textbf{83.71}                   & 83.39                 & \textbf{67.61}                   \\
    \bottomrule
    \end{tabular}
    ~
    \label{table:acc:newdata}
\end{table*}

\textbf{Result.} We first compare our approach with existing certified defense approaches in terms of clean accuracy and certified accuracy. As shown in Table~\ref{table:acc:baselines} and Table~\ref{table:acc:newdata}, our approach is able to obviously outperform the SOTA techniques with a clean accuracy of 81.8\% and a certified accuracy of 67.1\% on the challenging ImageNet with 32$\times$32-pixel adversarial patches. Due to the design of \sys, its clean accuracy remains the same as the base ViT model, which can be even further improved by using better base models. 

\footnotetext[3]{We are unable to find corresponding clean accuracy with respect to the best certified accuracy claimed by the authors}

The results of detection-based approaches (including ours) are not directly comparable with the recovery-based approaches because they are designed for different goals. However, we notice that our approach was able to achieve a much higher certified accuracy than recovery-based methods, so it may be more practical to use in the real world. The main reason is that the certified detection is based on voting over predictions with a small region excluded, which can still provide sufficient global information, rather than in recovery-based approaches where each voter is based on a small local patch.

{\responserefa{}As compared to other SOTA certified detection approaches, \sys could achieve higher certified accuracy on all four datasets. The results of clean accuracy are similar except for MR+ on Food-101, which means ViT-base performs slightly worse than ResNet-50. {\responserefb{} Notice that here we train both models with the same configuration (randomly occluded patches in the same size), to be relatively fair for comparison. The results indicate that the training configuration might not be optimal for ViT, which could still have some room for further improvement. Even though, under such a non-optimized configuration, the experiment results still confirm the effectiveness of our method in terms of certified accuracy.} The result shows that the superior certified accuracy of \sys comes not only from the better base performance of ViT but is also a result of the combination of {\sys}'s defense mechanism and ViT's robustness against the absent patch.  }


\begin{tcolorbox}[size=title]
	{\textbf{Answer to RQ1:}} 
\sys outperforms existing certified recovery methods by a large margin and also achieves higher clean accuracy and certified accuracy compared with other certified detection methods in most settings.
\end{tcolorbox}

\subsection{RQ2: Performance under Small {\responserefa{}ROI}}

\begin{figure}[tbp]
    \vspace{10mm}
    \centering
    \begin{subfigure}[b]{0.45\linewidth}
        \centering
        \includegraphics[width=1.5in]{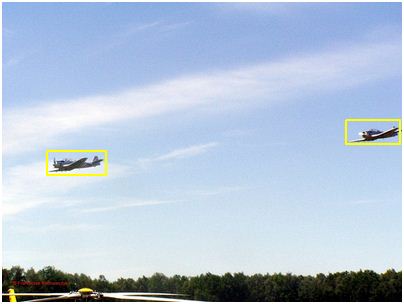}
        \caption{ImageNet with bounding box}
        \label{fig:evaluation:part_example_bounding}
    \end{subfigure}
    ~
    \begin{subfigure}[b]{0.45\linewidth}
        \centering
        \includegraphics[width=1.5in]{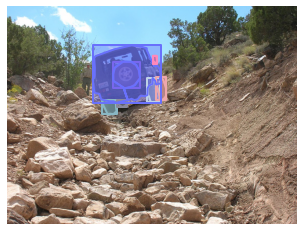}
        \caption{PartImageNet}
        \label{fig:evaluation:part_example_part}
    \end{subfigure}
    \caption{Illustration of the annotation}
\end{figure}

\textbf{Experimental Settings.} In this research question, we want to evaluate the performance of  \sys under small {\responserefa{}ROI}, as compared with the certified recovery methods mentioned in section~\ref{sec:motivation}. {\responserefa{}The adversarial patch size we aim to certify in this RQ is 2.4\%, which is $5 \times 5$ patch in a $32 \times 32$ image. We choose this patch size as it is the default setting in certified recovery work. } We first perform rescaling for each image in the CIFAR-10 dataset so the {\responserefa{}ROI} can be controlled. Then we retrain the DS-ResNet on the rescaled image following the default setting in De-randomized Smoothing~\cite{levine2020randomized} and PatchGuard~\cite{xiang2020patchguard}. To further validate the \sys can perform well on small {\responserefa{}ROI}, we additionally design two experiments on the ImageNet visual object detection~\cite{su2012crowdsourcing} and PartImageNet dataset \cite{he2021partImageNet}. The former includes bounding box annotations for the target object (as shown in Fig~\ref{fig:evaluation:part_example_bounding}) in the image and could serve as an indicator of how large the {\responserefa{}Region of interest(ROI)} is. We compute the {\responserefa{}ROI} by dividing the bounding box's area by the whole image's area. The latter includes more fine-grained per-pixel part annotations, and each object is partitioned into smaller parts (as shown in Fig~\ref{fig:evaluation:part_example_part}). As current certified recovery methods usually rely on local features for inference and verification, the area of object parts could also stand for {\responserefa{}ROI}. The {\responserefa{}ROI} of PartImageNet is obtained by dividing the area of the largest part by the area of the whole image. {\responserefb{} The model architectures we used for all the methods here remain the same with RQ1.}

\begin{figure*}[tbp]
\centering
    \begin{minipage}{1.0\linewidth}
        \captionsetup{width=1.0\linewidth}
        \begin{subfigure}[b]{0.48\linewidth}
          \centering
          \includegraphics[width=2.18in]{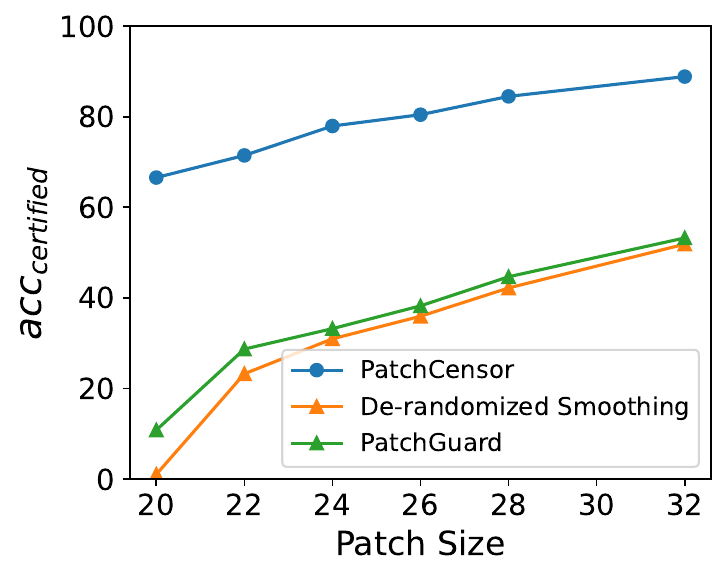}
        \end{subfigure}
        \begin{subfigure}[b]{0.48\linewidth}
          \centering
          \includegraphics[width=2.18in]{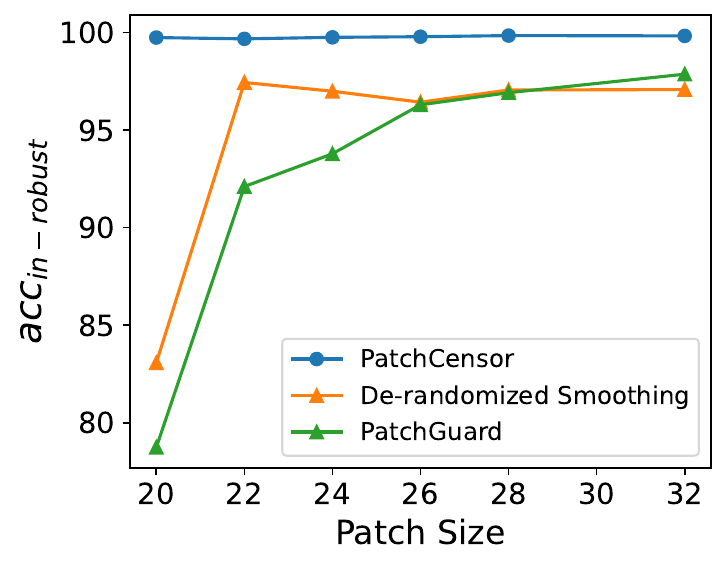}
        \end{subfigure}
        \caption{Certified accuracy and accuracy in the robust domain of \sys, PatchGuard and De-randomized Smoothing on rescaled CIFAR-10. }
        \label{fig:evaluation:resize_accuracy}
    \end{minipage}
\end{figure*}

\textbf{Result.} For the rescaled CIFAR-10 experiment, detailed result is shown at Table~\ref{table:evaluation:resize}. We also plot the changing trend of $\accCertified$ and $\accInTrust$ for \sys and other two techniques in Figure~\ref{fig:evaluation:resize_accuracy}.

One interesting observation is that certified recovery techniques (both the PatchGuard and De-randomized Smoothing) experience a quick drop for both $\rTrust$ and $\accInTrust$ when the image size scales from 32$\times$32 to 20$\times$20. The $\rTrust$ of De-randomized Smoothing even drops to nearly 0 (1.24\%) with the rescaling size 20, unable to give certified prediction anymore. The reason behind this is the voting mechanism discussed in section~\ref{sec:motivation}.
While \sys is able to remain a high certified accuracy even with very small ROI.

\begin{table*}[]
\centering
\scriptsize
\captionsetup{width=.8\linewidth}
\caption{The clean accuracy, certified accuracy, and accuracy in trust domain achieved by \sys (PC), PatchGuard (PG) and De-randomized Smoothing (DS) for 2.4\%-pixel adversarial patch attack.}
\resizebox{1\columnwidth}{!}{ 
        \begin{tabular}{c|ccc|ccc|ccc}
    \toprule        
        \multirow{2}{*}{Rescaling size} & \multicolumn{3}{c|}{$\accClean$} & \multicolumn{3}{c|}{$\accCertified$} & \multicolumn{3}{c}{$\accInTrust$}  \\
                                          & PC               & PG              & DS                 & PC                & PG               & DS              & PC                  & PG               & DS                 \\
    \midrule
        32 (original size)                 & \textbf{98.8}            & 83.35           & 83.35              & \textbf{88.85}            & 53.23            & 51.85           & \textbf{99.83}               & 97.87               & 97.08             \\
    \midrule
        28                 & \textbf{98.3}            & 80.80           & 80.29             & \textbf{84.48}             & 44.66            & 42.19           & \textbf{99.85}               & 96.92               & 97.06              \\
        26                & \textbf{97.81}            & 77.47          & 77.65              & \textbf{80.44}             & 38.24            & 35.94           & \textbf{99.79}               & 96.3              &96.43              \\
        24                & \textbf{97.79}            & 77.05           & 76.85             & \textbf{77.94}             & 33.21            & 30.96           & \textbf{99.76}               & 93.79               &96.99              \\
        22               & \textbf{97.14}            & 76.26           & 75.63              & \textbf{71.45}             & 28.70            & 23.24           & \textbf{99.68}               & 92.11              & 97.44              \\
        20               & \textbf{96.67}            & 69.38           & 67.35              & \textbf{66.56}             & 10.81            & 1.03           & \textbf{99.75}               & 78.73               & 83.06              \\
    \bottomrule
    \end{tabular}
    ~
}

    \label{table:evaluation:resize}
\end{table*}

\begin{figure*}[tbp]
\centering
    \begin{minipage}{1.0\linewidth}
        \captionsetup{width=1.0\linewidth}
        \begin{subfigure}[b]{0.48\linewidth}
          \centering
          \includegraphics[width=2.18in]{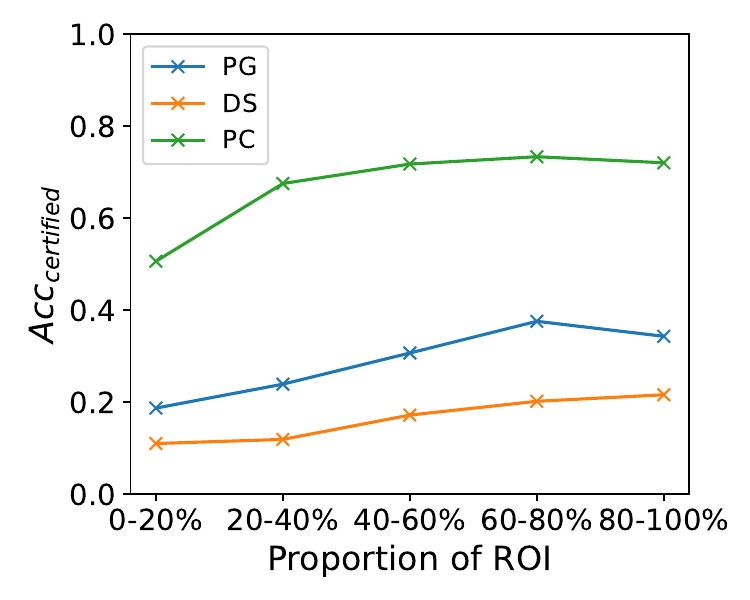}
        \end{subfigure}
        \begin{subfigure}[b]{0.48\linewidth}
          \centering
          \includegraphics[width=2.18in]{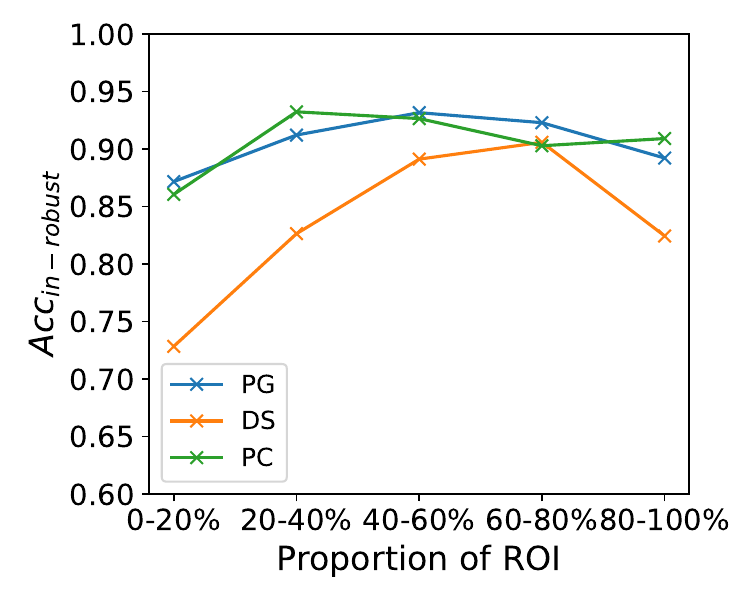}
        \end{subfigure}
        \caption{Certified accuracy and accuracy in the robust domain of \sys, PatchGuard and De-randomized Smoothing on ImageNet}
        \label{fig:evaluation:ImageNet_bounding}
    \end{minipage}
\end{figure*}

\begin{figure*}[tbp]
\centering
    \begin{minipage}{1.0\linewidth}
        \captionsetup{width=1.0\linewidth}
        \begin{subfigure}[b]{0.48\linewidth}
          \centering
          \includegraphics[width=2.18in]{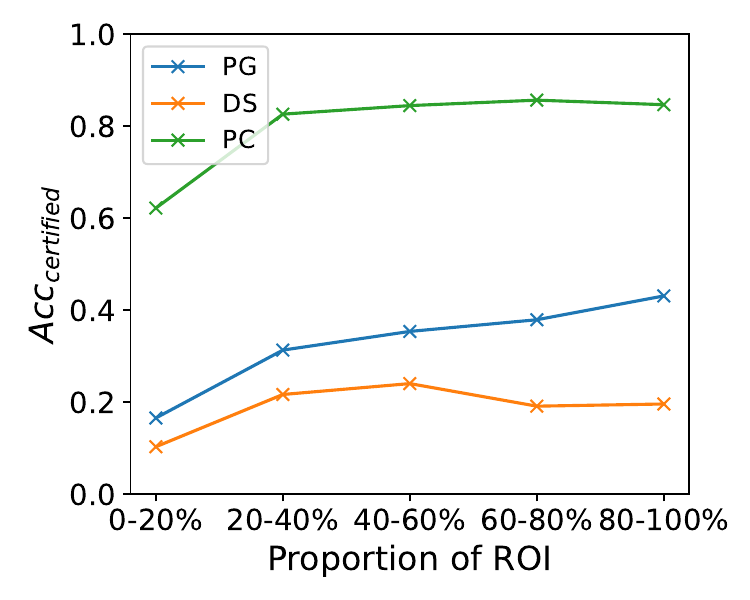}
        \end{subfigure}
        \begin{subfigure}[b]{0.48\linewidth}
          \centering
          \includegraphics[width=2.18in]{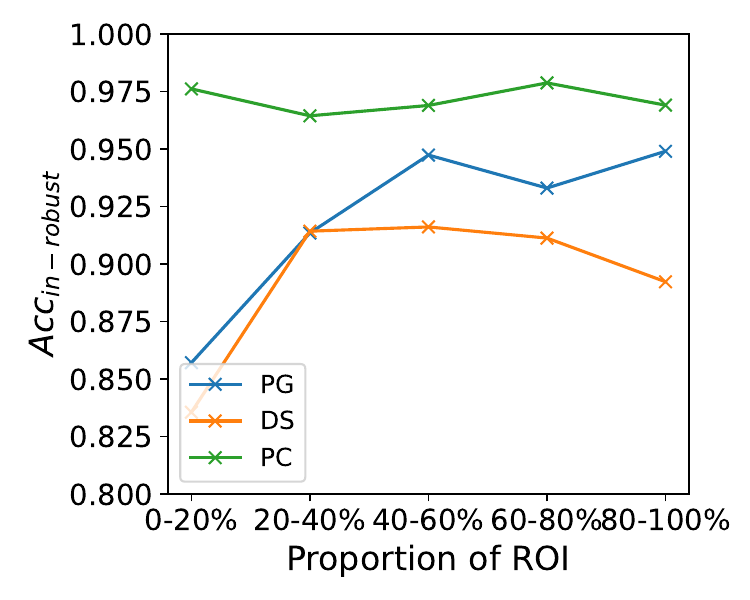}
        \end{subfigure}
        \caption{Certified accuracy and accuracy in the robust domain of \sys, PatchGuard and De-randomized Smoothing on PartImageNet}
        \label{fig:evaluation:partImageNet}
    \end{minipage}
\end{figure*}

For experiments on ImageNet object detection and PartImageNet, we first sort the image according to {\responserefa{}ROI} in ascending order and compute certified accuracy in 5-quantiles. The result for the former is shown in Figure~\ref{fig:evaluation:ImageNet_bounding}, and the result for the latter is shown in Figure~\ref{fig:evaluation:partImageNet}. It can be observed that \sys can still have 51.0\% certified accuracy on ImageNet object detection and 72.6\% on PartImageNet in the 0-20\% region, while PatchGuard and De-randomized Smoothing can only yield 18.7\% (24.2\%),  and 11.0\% (16.2\%) certified accuracy, respectively.

It is worth noticing that $\accInTrust$ of PatchCensor has some overlaps across different {\responserefa{}ROI} on ImageNet object detection and PartImageNet. This may be due to the fact that pre-trained models on ImageNet may utilize other features in the image, such as texture instead of shape, as pointed out by Geirhos \etal~\cite{geirhos2018ImageNet}. In such cases, using only local features may be enough. However, this can be a kind of overfitting. Our controlled experiments on CIFAR-10 have no such issue. Despite this, extra information such as texture could provide performance gain for certified recovery, and we will discuss more in Section~\ref{sec:discussion}.

\begin{tcolorbox}[size=title]
	{\textbf{Answer to RQ2:}} 
\sys is able to achieve moderate performance when {\responserefa{}ROI} is small. It can even maintain relatively good performance under small {\responserefa{}ROI} at more complex and challenging ImageNet dataset.
\end{tcolorbox}

\subsection{RQ3: Performance under Strong Adversaries}

\textbf{Experimental Settings.} In this research question, we evaluate the certification performance of \sys and MR+ with large adversarial patches whose size is up to 25.0\%. We ignore certified recovery work because they only focus on small patches and are unable to handle large patches. On the one hand, the certification technique yields few certified instances once the patch size becomes large. On the other hand, it is much harder to train the model under such a setting. However, our defense and other SOTA certified detection work focus mainly on masking strategy. It is possible to certify instances even when the adversarial patches are large in size. Being able to certify inputs with large patch sizes would also make it possible to detect abnormal scenarios other than the adversarial setting, such as natural perturbation~\cite{gu2022evaluating}, occlusion in object detection~\cite{zhang2018deepvoting} and open-set detection~\cite{acsintoae2022ubnormal}. As natural objects are often larger than commonly evaluated 2.0\% patch size. {\responserefa{}In the experiments, we select CIFAR-10, GTSRB, Food-101 and ImageNet as the subject datasets for evaluation. For the models, ViT-B/16 ({\responserefb{}the same architecture in RQ1, RQ2}) and ResNet50 pre-trained on ImageNet are selected for PatchCensor and MR++ accordingly. Additionally, as we target a large adversarial patch size in this RQ, fine-tuning on randomly dropped patches can potentially improve the certified accuracy for both methods. As a result, we report the performance with/without such fine-tuning to provide a more comprehensive analysis of both methods. This leads to eight groups of experiments in total as shown below.} 

{\responserefa{}
    \textbf{Result.} We tested the defense effectiveness of \sys and MR+ under different adversarial patch sizes on four datasets, with a total of 112 configurations. The results for for CIFAR-10 are summarized in Table~\ref{table:acc:patch_sizes_cifar10} and Table~\ref{table:acc:patch_sizes_cifar10_ft}, results for GTSRB are summarized in Table~\ref{table:acc:patch_sizes_GTSRB} and Table~\ref{table:acc:patch_sizes_GTSRB_ft}, results for Food-101 are summarized in Table~\ref{table:acc:patch_sizes_Food101} and Table~\ref{table:acc:patch_sizes_Food101_ft}, and the results for ImageNet are summarized in Table~\ref{table:acc:patch_sizes_ImageNet} and Table~\ref{table:acc:patch_sizes_ImageNet_ft}.

    For all four datasets, \sys achieves better performance on the most important certified accuracy metric than MR+ in most of the settings. The maximum difference is 40.73\% on CIFAR-10 without fine-tuning, and the minimum difference is 2.57\% on GTSRB without fine-tuning. The only exceptions are on the CIFAR-10 dataset with fine-tuning, where MR+ surpasses \sys when the patch size is larger or equal to 18.4\%. From the result, we infer that CNN, specifically fine-tuned for patch drop, can be more robust than ViT on a simper dataset such as CIFAR-10. 

    As for the clean accuracy, \sys is higher than MR+ on nearly all settings (ranging from 1.18\% on ImageNet to 8.31\% on fine-tuned GTSRB) except for Food-101 with fine-tuning. Still, even under such a circumstance, \sys has higher certified accuracy than MR+ across all adversarial patches on Food-101. 
    

    Interestingly, both methods have the worst performance on the GTSRB dataset when the adversarial patch size is large. It can be surprising that the certified accuracy of GTSRB is even lower than that of ImageNet, which is believed to be far more complex than GTSRB. Also, fine-tuning plays an important role in the performance,  giving 26.87\% improvement for \sys and 19.02\% improvement for MR+ on certified accuracy. The reason may be that, compared with ImageNet,  the target object in GTSRB occupies most of the image. It is more likely that the dropped patch is inside the target object and thus has more effect on the data distribution, interfering with the model's prediction. When the size of the dropped patch becomes large, important details may be missed, causing a sharp decrease in performance. As such, we suggest fine-tuning the models under a similar situation.
    
    
    For other subject datasets we evaluated besides the GTSRB dataset, the fine-tuning process does not influence the performance of \sys too much. The average increase after fine-tuning for \sys in certified accuracy across all patches is 5.39\%, -0.73 \%, and 4.44\% for CIFAR-10,  Food-101, and ImageNet, respectively. However, MR+ relies heavily on fine-tuning. The corresponding increase in certified accuracy is 33.23\%, 19.02\%, and 4.72\% for CIFAR-10, Food-101, and ImageNet. Furthermore, the certified accuracy of \sys without fine-tuning is higher than that of MR+ with fine-tuning on 16 of 21 cases on these three datasets. We can thus conclude that \sys can yield moderate performance on datasets similar to these three, even without fine-tuning. This means that the pre-trained models can be directly combined with \sys to increase its patch robustness. 
    
}

Meanwhile, it is interesting to notice that the accuracy of \sys in the trust domain ($acc_{\text{in-trust}}$) remained high (even slightly increases) under larger adversarial patch sizes for all datasets. This means that \sys can retain high usability under strong adversaries - even though \sys may raise more warnings (\ie report more unverifiable input images) when defending against stronger adversaries, it can still promise a high accuracy when it gives verified predictions.

\begin{table}[]
\centering
\scriptsize
\responserefa{}
\caption{{\responserefa{}Clean accuracy \& certified accuracy achieved by \sys and MR+ on \textbf{CIFAR-10} under different adversarial patch sizes \textbf{without fine-tuning}}}

    \resizebox{1\columnwidth}{!}{ 
        \begin{tabular}{c|cc|cc|cc|cc}
    \toprule        
        \multirow{2}{*}{Max Adv Patch Size} & \multicolumn{2}{c|}{$\accClean$} & \multicolumn{2}{c|}{$\accCertified$} & \multicolumn{2}{c|}{$\rTrust$} & \multicolumn{2}{c}{$\accInTrust$} \\
                                          & PC               & MR+              & PC                 & MR+                & PC               & MR+              & PC                  & MR+                 \\
    \midrule
        0.5\%                 & \textbf{98.74}            & 96.80           & \textbf{94.57}              & 79.48            & \textbf{94.89}            & 79.84           & \textbf{99.66}               & 99.55              \\
        2.0\%                 & \textbf{98.74}            & 96.81           & \textbf{90.13}              & 68.52            & \textbf{90.29}            & 68.78           & \textbf{99.82}               & 99.62            \\
        4.6\%                 & \textbf{98.74}            & 96.80           & \textbf{84.12}              & 54.90             & \textbf{84.24}            & 55.07           & \textbf{99.86}              & 99.69              \\
        8.2\%                 & \textbf{98.74}            & 96.79           & \textbf{76.20}              & 41.37             & \textbf{76.28}            & 41.53           & \textbf{99.90}              & 99.61              \\
        12.8\%                & \textbf{98.74}            & 96.81            & \textbf{67.41}              & 31.32             & \textbf{67.46}            & 31.53           & \textbf{99.93}               & 99.33              \\
        18.4\%                & \textbf{98.74}            & 96.81           & \textbf{58.90}             & 23.79             & \textbf{58.97}            & 24.01           & \textbf{99.88}               & 99.08              \\
        25.0\%                & \textbf{98.74}            & 96.79           & \textbf{48.52}             & 18.50             & \textbf{48.57}            & 18.63           & \textbf{99.90}             & 99.30              \\
    \bottomrule
    \end{tabular}
    ~
}
    \label{table:acc:patch_sizes_cifar10}
\end{table}

\begin{table}[]
\centering
\scriptsize
\responserefa{}
\caption{{\responserefa{}Clean accuracy \& certified accuracy achieved by \sys and MR+ on \textbf{CIFAR-10} under different adversarial patch sizes \textbf{with fine-tuning}}}

    \resizebox{1\columnwidth}{!}{ 
        \begin{tabular}{c|cc|cc|cc|cc}
    \toprule        
        \multirow{2}{*}{Max Adv Patch Size} & \multicolumn{2}{c|}{$\accClean$} & \multicolumn{2}{c|}{$\accCertified$} & \multicolumn{2}{c|}{$\rTrust$} & \multicolumn{2}{c}{$\accInTrust$} \\
                                          & PC               & MR+              & PC                 & MR+                & PC               & MR+              & PC                  & MR+                 \\
    \midrule
        0.5\%                 & \textbf{98.77}            & 97.61           & \textbf{95.70}              & 90.59            & \textbf{96.05}            & 90.97           & \textbf{99.64}               & 99.58              \\
        2.0\%                 & \textbf{98.80}            & 97.73          & \textbf{92.84}              & 86.77            & \textbf{93.06}            & 87.00           & \textbf{99.76}               & 99.74            \\
        4.6\%                 & \textbf{98.81}            & 97.70           & \textbf{88.29}              & 83.30             & \textbf{88.42}            & 83.46           & \textbf{99.85}              & 99.81              \\
        8.2\%                 & \textbf{98.77}            & 97.51           & \textbf{82.90}              & 79.78             & \textbf{83.04}            & 79.82           & \textbf{99.83}              & 99.82              \\
        12.8\%                & \textbf{98.76}            & 97.35           & \textbf{75.52}              & 74.91             & \textbf{75.99}            & 75.04           & \textbf{99.91}               & 99.83              \\
        18.4\%                & \textbf{98.76}            & 96.90           & 66.68             & \textbf{71.03}             & 66.77            & \textbf{71.19}           & \textbf{99.87}               & 99.78              \\
        25.0\%                & \textbf{98.75}            & 95.57           & 55.89             & \textbf{65.60}             & 55.95            & \textbf{65.77}           & \textbf{99.89}             & 99.74              \\
    \bottomrule
    \end{tabular}
    ~
}
    \label{table:acc:patch_sizes_cifar10_ft}
\end{table}

\begin{table}[]
\centering
\scriptsize
\responserefa{}
\caption{{\responserefa{}Clean accuracy \& certified accuracy achieved by \sys and MR+ on \textbf{GTSRB} under different adversarial patch sizes \textbf{without fine-tuning}}}

    \resizebox{1\columnwidth}{!}{ 
        \begin{tabular}{c|cc|cc|cc|cc}
    \toprule        
        \multirow{2}{*}{Max Adv Patch Size} & \multicolumn{2}{c|}{$\accClean$} & \multicolumn{2}{c|}{$\accCertified$} & \multicolumn{2}{c|}{$\rTrust$} & \multicolumn{2}{c}{$\accInTrust$} \\
                                          & PC               & MR+              & PC                 & MR+                & PC               & MR+              & PC                  & MR+                 \\
    \midrule
        0.5\%                 & \textbf{97.43}            & 92.49           & \textbf{70.26}              & 36.13            & \textbf{71.01}            & 36.32           & 98.94               & \textbf{99.48}              \\
        2.0\%                 & \textbf{97.43}            & 92.48          & \textbf{40.89}              & 22.04            & \textbf{41.31}            & 22.06           & 98.97               & \textbf{99.93}           \\
        4.6\%                 & \textbf{97.43}           & 92.48          & \textbf{26.04}              & 11.94             & \textbf{26.08}            & 11.94          & 99.85              & \textbf{100}              \\
        8.2\%                 & \textbf{97.43}             & 92.49           & \textbf{18.39}              & 7.19             & \textbf{18.42}            & 7.21           & \textbf{99.87}              & 99.78              \\
        12.8\%                & \textbf{97.43}            & 92.48           & \textbf{12.49}              & 5.65             & \textbf{12.50}            & 5.69           & \textbf{99.87}              & 99.30              \\
        18.4\%                & \textbf{97.43}            & 92.49           & \textbf{8.99}             & 4.73             & \textbf{9.02}            & 4.79           & \textbf{99.74}               & 98.68              \\
        25.0\%                & \textbf{97.43}            & 92.49           & \textbf{6.71}             & 4.14             & \textbf{6.73}            & 4.22           & \textbf{99.76}             & 98.12              \\
    \bottomrule
    \end{tabular}
    ~
}
    \label{table:acc:patch_sizes_GTSRB}
\end{table}

\begin{table}[]
\centering
\scriptsize
\responserefa{}
\caption{{\responserefa{}Clean accuracy \& certified accuracy achieved by \sys and MR+ on \textbf{GTSRB} under different adversarial patch sizes \textbf{with fine-tuning}}}

    \resizebox{1\columnwidth}{!}{ 
        \begin{tabular}{c|cc|cc|cc|cc}
    \toprule        
        \multirow{2}{*}{Max Adv Patch Size} & \multicolumn{2}{c|}{$\accClean$} & \multicolumn{2}{c|}{$\accCertified$} & \multicolumn{2}{c|}{$\rTrust$} & \multicolumn{2}{c}{$\accInTrust$} \\
                                          & PC               & MR+              & PC                 & MR+                & PC               & MR+              & PC                  & MR+                 \\
    \midrule
        0.5\%                 & \textbf{96.84}            & 91.31           & \textbf{83.10}              & 56.79            & \textbf{83.97}            & 58.42           & \textbf{98.96}               & 97.21            \\
        2.0\%                 & \textbf{97.09}            & 92.53          & \textbf{70.33}              & 48.35            & \textbf{71.21}            & 49.43           & \textbf{98.77}               & 97.81          \\
        4.6\%                 & \textbf{96.88}           & 91.09          & \textbf{61.29}              & 36.48             & \textbf{61.88}            & 37.61          & \textbf{99.05}              & 97.01             \\
        8.2\%                 & \textbf{97.65}             & 91.23           & \textbf{49.29}              & 27.78             & \textbf{49.89}            & 28.35           & \textbf{98.79}              & 97.96              \\
        12.8\%                & \textbf{97.32}            & 89.96           & \textbf{43.93}              & 21.77             & \textbf{44.36}            & 22.33           & \textbf{99.02}              & 97.48              \\
        18.4\%                & \textbf{96.52}            & 90.15           & \textbf{35.61}             & 18.38             & \textbf{36.08}            & 18.65           & \textbf{98.71}               & 98.51              \\
        25.0\%                & \textbf{96.33}            & 88.02           & \textbf{28.31}             & 15.39             & \textbf{28.86}            & 15.86           & \textbf{98.08}             & 97.05              \\
    \bottomrule
    \end{tabular}
    ~
}
    \label{table:acc:patch_sizes_GTSRB_ft}
\end{table}

\begin{table}[]
\centering
\scriptsize
\responserefa{}
\caption{{\responserefa{}Clean accuracy \& certified accuracy achieved by \sys and MR+ on \textbf{Food-101} under different adversarial patch sizes \textbf{without fine-tuning}}}

    \resizebox{1\columnwidth}{!}{ 
        \begin{tabular}{c|cc|cc|cc|cc}
    \toprule        
        \multirow{2}{*}{Max Adv Patch Size} & \multicolumn{2}{c|}{$\accClean$} & \multicolumn{2}{c|}{$\accCertified$} & \multicolumn{2}{c|}{$\rTrust$} & \multicolumn{2}{c}{$\accInTrust$} \\
                                          & PC               & MR+              & PC                 & MR+                & PC               & MR+              & PC                  & MR+                 \\
    \midrule
        0.5\%                 & \textbf{86.46}            & 84.45           & \textbf{77.08}              & 66.28            & \textbf{81.04}            & 68.62           & 95.11               & \textbf{96.59}              \\
        2.0\%                 & \textbf{86.46}            & 84.46         & \textbf{72.30}              & 60.38            & \textbf{74.91}            & 61.85           & 96.52               & \textbf{97.63}           \\
        4.6\%                 & \textbf{86.46}           & 84.46           & \textbf{66.60}              & 53.36             & \textbf{68.26}            & 54.38          & 97.56              & \textbf{98.12}              \\
        8.2\%                 & \textbf{86.46}             & 84.46            & \textbf{60.02}              & 45.89             & \textbf{61.15}            & 46.55           & 98.15              & \textbf{98.58}               \\
        12.8\%                & \textbf{86.46}            & 84.46            & \textbf{52.82}              & 38.71             & \textbf{53.61}            & 39.17           & 98.52              & \textbf{98.82}               \\
        18.4\%                & \textbf{86.46}            & 84.46            & \textbf{44.93}             & 31.71             & \textbf{45.52}            & 32.05           & 98.70               & \textbf{98.94}               \\
        25.0\%                & \textbf{86.46}            & 84.46            & \textbf{36.61}             & 24.74             & \textbf{37.09}            & 25.03           & 98.73             & \textbf{98.83}              \\
    \bottomrule
    \end{tabular}
    ~
}
    \label{table:acc:patch_sizes_Food101}
\end{table}

\begin{table}[]
\centering
\scriptsize
\responserefa{}
\caption{{\responserefa{}Clean accuracy \& certified accuracy achieved by \sys and MR+ on \textbf{Food-101} under different adversarial patch sizes \textbf{with fine-tuning}}}

    \resizebox{1\columnwidth}{!}{ 
        \begin{tabular}{c|cc|cc|cc|cc}
    \toprule        
        \multirow{2}{*}{Max Adv Patch Size} & \multicolumn{2}{c|}{$\accClean$} & \multicolumn{2}{c|}{$\accCertified$} & \multicolumn{2}{c|}{$\rTrust$} & \multicolumn{2}{c}{$\accInTrust$} \\
                                          & PC               & MR+              & PC                 & MR+                & PC               & MR+              & PC                  & MR+                 \\
    \midrule
        0.5\%                 & 83.28            & \textbf{84.77}          & \textbf{72.56}              & 70.57            & \textbf{77.66}            & 73.67           & 93.44               & \textbf{95.79}              \\
        2.0\%                 & 83.39            & \textbf{84.68}         & \textbf{67.61}              & 64.80            & \textbf{70.93}            & 66.86           & 95.32               & \textbf{96.91}           \\
        4.6\%                 & 84.02           & \textbf{84.82}           & \textbf{63.35}              & 59.16             & \textbf{65.42}            & 60.40          & 96.84              & \textbf{97.95}              \\
        8.2\%                 & 84.80             & \textbf{84.47}            & \textbf{58.65}              & 53.24             & \textbf{59.96}            & 54.18           & 97.81              & \textbf{98.27}               \\
        12.8\%                & 84.90            & \textbf{85.06}            & \textbf{53.53}              & 48.58             & \textbf{54.42}            & 49.21           & 98.37              & \textbf{98.73}               \\
        18.4\%                & 84.97            & \textbf{85.30}            & \textbf{47.22}             & 44.33             & \textbf{47.92}            & 44.83           & 98.55               & \textbf{98.88}               \\
        25.0\%                & \textbf{85.62}            & 85.33            & \textbf{42.32}             & 38.57             & \textbf{42.81}            & 39.02           & \textbf{98.85}             & 98.84              \\
    \bottomrule
    \end{tabular}
    ~
}
    \label{table:acc:patch_sizes_Food101_ft}
\end{table}

    \begin{table}[]
    \centering
    \scriptsize
    \responserefa{}
    \caption{{\responserefa{}Clean accuracy \& certified accuracy achieved by \sys and MR+ on \textbf{ImageNet} under different adversarial patch sizes \textbf{without fine-tuning}}}
    
        \resizebox{1\columnwidth}{!}{ 
            \begin{tabular}{c|cc|cc|cc|cc}
        \toprule        
            \multirow{2}{*}{Max Adv Patch Size} & \multicolumn{2}{c|}{$\accClean$} & \multicolumn{2}{c|}{$\accCertified$} & \multicolumn{2}{c|}{$\rTrust$} & \multicolumn{2}{c}{$\accInTrust$} \\
                                              & PC               & MR+              & PC                 & MR+                & PC               & MR+              & PC                  & MR+                 \\
        \midrule
            0.5\%                 & \textbf{81.8}            & 80.62           & \textbf{72.0}              & 62.68             & \textbf{80.3}            & 68.27           & 89.73               & \textbf{91.81}              \\
            2.0\%                 & \textbf{81.8}            & 80.62          & \textbf{67.2}              & 56.02            & \textbf{73.5}            & 60.24           & 91.35               & \textbf{92.99}              \\
            4.6\%                 & \textbf{81.8}            & 80.62           & \textbf{61.9}              & 49.52             & \textbf{67.0}            & 52.75           & 92.44               & \textbf{93.88}              \\
            8.2\%                 & \textbf{81.8}            & 80.62           & \textbf{56.4}              & 42.94             & \textbf{60.4}            & 45.46           & 93.29               & \textbf{94.46}              \\
            12.8\%                & \textbf{81.8}            & 80.62           & \textbf{50.5}              & 36.16             & \textbf{53.7}            & 38.22           & 93.96               & \textbf{94.62}              \\
            18.4\%                & \textbf{81.8}            & 80.62           & \textbf{44.1}              & 30.05             & \textbf{46.7}            & 31.70           & 94.57               & \textbf{94.78}              \\
            25.0\%                & \textbf{81.8}            & 80.62           & \textbf{37.1}              & 24.07             & \textbf{39.1}            & 25.37           & 94.80              & \textbf{94.88}              \\
        \bottomrule
        \end{tabular}
        ~
    }
        \label{table:acc:patch_sizes_ImageNet}
    \end{table}

\begin{table}[]
\centering
\scriptsize
\responserefa{}
\caption{{\responserefa{}Clean accuracy \& certified accuracy achieved by \sys and MR+ on \textbf{ImageNet} under different adversarial patch sizes \textbf{with fine-tuning}}}

    \resizebox{1\columnwidth}{!}{ 
        \begin{tabular}{c|cc|cc|cc|cc}
    \toprule        
        \multirow{2}{*}{Max Adv Patch Size} & \multicolumn{2}{c|}{$\accClean$} & \multicolumn{2}{c|}{$\accCertified$} & \multicolumn{2}{c|}{$\rTrust$} & \multicolumn{2}{c}{$\accInTrust$} \\
                                          & PC               & MR+              & PC                 & MR+                & PC               & MR+              & PC                  & MR+                 \\
    \midrule
        0.5\%                 & \textbf{82.70}            & 75.59           & \textbf{73.67}              & 61.84            & \textbf{81.88}            & 70.99           & \textbf{89.98}               & 87.12              \\
        2.0\%                 & \textbf{82.73}            & 75.51           & \textbf{69.41}              & 56.31             & \textbf{75.92}            & 63.05           & \textbf{91.43}               & 89.32              \\
        4.6\%                 & \textbf{82.67}            & 75.49           & \textbf{64.97}              & 51.23             & \textbf{70.27}            & 56.56           & \textbf{92.45}               & 90.57              \\
        8.2\%                 & \textbf{82.66}            & 76.34           & \textbf{60.54}              & 48.17             & \textbf{64.92}            & 52.55           & \textbf{93.26}               & 91.66              \\
        12.8\%                & \textbf{82.55}            & 76.20           & \textbf{55.38}              & 43.39             & \textbf{59.08}            & 46.99           & \textbf{93.73}               & 92.35              \\
        18.4\%                & \textbf{82.51}            & 75.81           & \textbf{50.66}              & 38.77             & \textbf{53.76}            & 41.63           & \textbf{94.23}               & 93.12              \\
        25.0\%                & \textbf{82.49}            & 75.50           & \textbf{45.62}              & 34.79             & \textbf{48.24}            & 37.34           & \textbf{94.56}               & 93.19              \\
    \bottomrule
    \end{tabular}
    ~
}
    \label{table:acc:patch_sizes_ImageNet_ft}
\end{table}

\begin{tcolorbox}[size=title]
	{\textbf{Answer to RQ3:}} 
{\responserefa{}\sys outperforms MR+ in most scenarios on CIFAR-10, GTSRB, Food-101 and ImageNet datasets.} It also retains high accuracy in the robust domain ($\accInTrust$) even with large adversarial patches. Being capable of detecting large abnormal patches makes it possible to use \sys in more complex natural perturbation settings.
\end{tcolorbox}

\subsection{RQ4: Overhead of \sys}

\textbf{Experimental Settings.} In this research question, we investigate how much computation overhead \sys may incur. Typically, there are two kinds of overhead for DNN models when conducting the certification. One is overhead for retraining/fine-tuning the model for the designed property. The other is the certification latency. 

{\responserefa{}For training overhead, we have already provided results and analysis with/without fine-tuning on four datasets in RQ3. For CIFAR-10, Food-101, and ImageNet, \sys can yield moderate results without any fine-tuning efforts. }  Another particular feature of \sys regarding the training overhead we want to emphasize here is that for both fine-tuning and certification, \sys targets adversarial patches whose size is in a specific range (\ie for all patches whose size is smaller than a threshold). In contrast, MR and most other patch certification methods only target adversarial patches with only one size. 

\begin{figure*}[tbp]
    \centering
    \begin{minipage}{1.0\linewidth}
    \captionsetup{width=1.0\linewidth}
        \begin{subfigure}[b]{0.48\linewidth}
            \centering
            \includegraphics[width=2.2in]{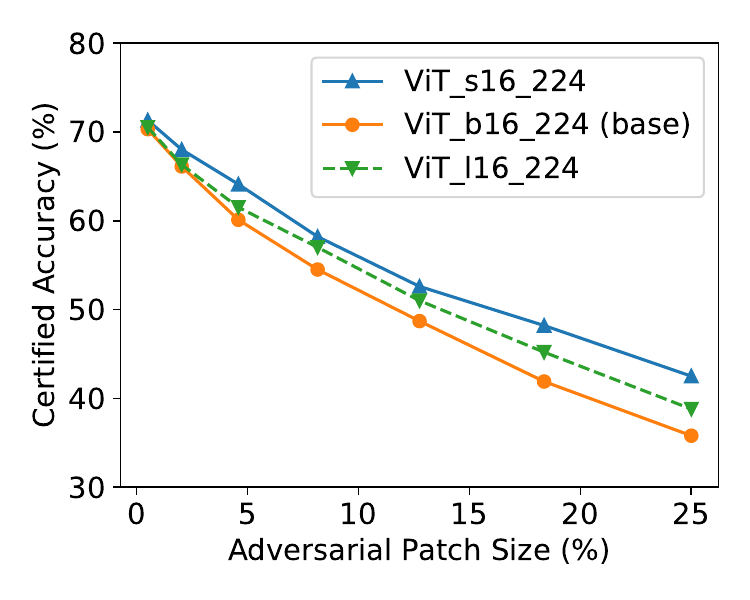}
        \end{subfigure}
        ~
        \begin{subfigure}[b]{0.48\linewidth}
            \centering
            \includegraphics[width=2.2in]{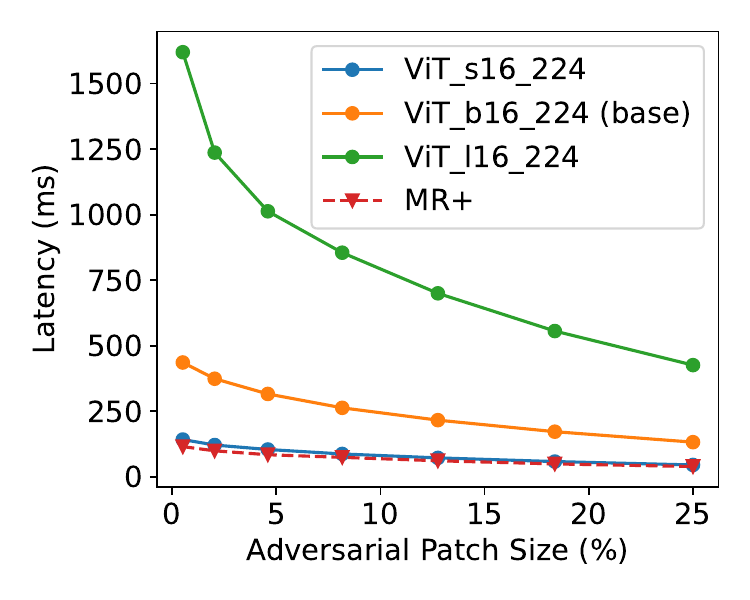}
        \end{subfigure}
    \caption{{\responserefa{}Certified accuracy and latency for ViT variants of different model sizes}}
    \label{fig:evaluation:latency_model_size}
    \end{minipage}
\end{figure*}
\begin{figure*}[tbp]
    \centering
    \begin{minipage}{1.0\linewidth}
    \captionsetup{width=1.0\linewidth}
        \begin{subfigure}[b]{0.48\linewidth}
            \centering
            \includegraphics[width=2.2in] {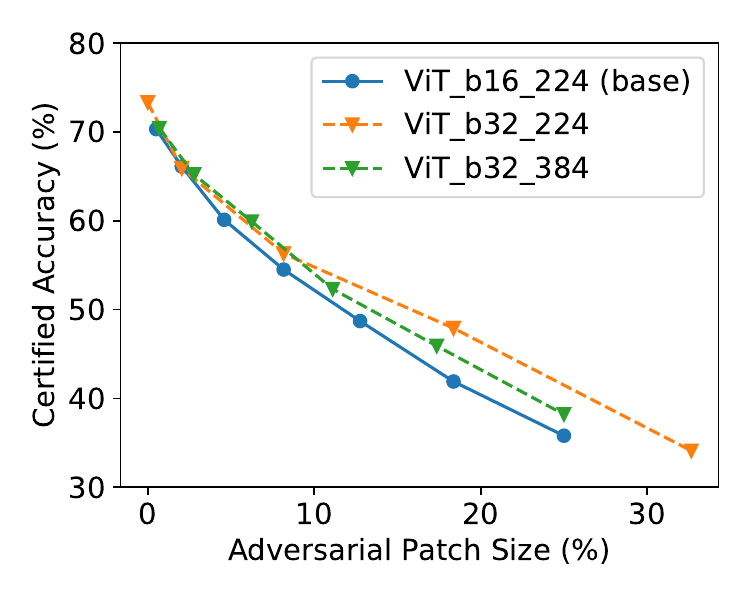}
        \end{subfigure}
        ~
        \begin{subfigure}[b]{0.48\linewidth}
            \centering
            \includegraphics[width=2.2in]{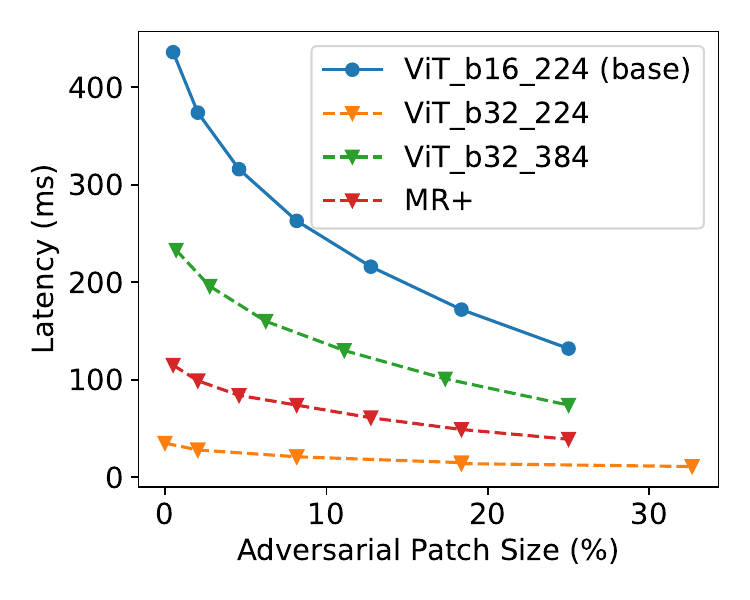}
        \end{subfigure}
    \caption{{\responserefa{}Certified accuracy and latency for ViT variants of different input and patch sizes}}
    \label{fig:evaluation:latency_input_size}
    \end{minipage}
\end{figure*}

For verification latency, we measured the latency of \sys to verify an input image against different adversarial patch sizes. Here we compare our technique with the previous SOTA MR+, as mentioned in the design goal of RQ4.
\textbf{Results.}
{\responserefa{}
    We measure the verification latency of \sys to see whether it is acceptable in this RQ. As ViT has many variants, and the latency for them may have a high variance, we independently provide results on each of them. We choose Vit\_s16\_224 (small), ViT\_b16\_224 (base), and ViT\_l16\_224 (large) for investigating the influence of model size. We choose ViT\_b16\_224 (base), ViT\_b32\_224 (base), and ViT\_b32\_384 (base) for investigating the influence of patch and input size. Model details are shown in table~\ref{tab:latency_model}. We additionally provide the results on clean accuracy and certified accuracy evaluated on 1,000 randomly sampled images from ImageNet without fine-tuning for defense. We hope this can serve as a guide for deployment in practice. 

    The results of the performance regarding the model sizes are shown in figure~\ref{fig:evaluation:latency_model_size}. It is surprising that the smallest model achieves both the best certified accuracy and lowest latency (similar latency with MR+) among all the three variants. The reason may be that this smallest model does not overfit the dataset compared to the other two bigger models and is thus more robust towards masking. A similar result is also reported in~\cite{naseer2021intriguing}, where the smallest ViT has a high accuracy for random patch drop when the information loss is high. For the other two variants, stronger backbones produce slightly higher certified accuracy as we expected. 

    The results of the performance regarding the input and patch sizes are shown in figure~\ref{fig:evaluation:latency_input_size}. These three variants have similar certified accuracy with quite different latency. Given the same input size, a higher patch size for the base model will drastically lower the latency. The ViT\_b32\_224 achieves even lower latency than MR+. However, a smaller base patch will provide a more subtle control of the certified patch size. As \sys certifies the patch \textit{in range}, its certified accuracy is in the format of the step function, such as 0-16, 16-32, 32-48, ... for base patch equals to 16 and 0-32, 32-64, 64-96, ... for base patch equals to 32. If one wishes to certify a patch with a size 47, it is better to choose a small base patch (32-48) rather than a big base patch(32-64) because the former can yield a better certified accuracy. 

    The results demonstrate that \sys is better when using a smaller model with a large base patch. Under such a scenario, \sys can achieve lower latency than MR+ while providing higher certified accuracy on a complex dataset.

    Still, we want to emphasize that \sys is not restricted to any kind of ViT variant. Even with a large ViT with a small base patch, the higher latency incurred by \sys is also a meaningful tradeoff in some realistic scenarios because of the various benefits it provides, such as no need for fine-tuning and higher certified accuracy. \sys is also a flexible detection framework that can work parallel with other methods. For real-time scenarios where the latency constraint is tight, PatchCensor can be combined and play as a nice complement to other low-latency algorithms, where it focuses on those critical frames that need rigorous analysis. 
}

\begin{table}[]
\responserefa{}
\caption{{\responserefa{}Choosed ViT variants for latency measurement.} {\responserefb{} The base model is what we use for evaluation in RQ1, RQ2, and RQ3.}}
\label{tab:latency_model}
\begin{tabular}{ccccc}
\hline
\textbf{ViT Variants}    & \textbf{Number of Parameters} & \textbf{Clean Accuracy} & \textbf{Patch Size} & \textbf{Input size}\\ \hline
ViT\_s16\_224 & 22,050,664  & 80.5 \%   & 16    & 224\\ \hline
ViT\_b16\_224 (base)  & 86,567,656     & 80.9 \%    & 16    & 224\\ \hline
ViT\_l16\_224  & 304,326,632  & 82.2 \% & 16    & 224\\ \hline
ViT\_b32\_224 & 88,224,232    & 80.8 \% & 32    & 224\\ \hline
ViT\_b32\_384 & 88,297,192    & 80.8 \% & 32    & 384\\ \hline
\end{tabular}
\end{table}

\begin{tcolorbox}[size=title]
	{\textbf{Answer to RQ4:}} 
    \sys requires much less effort in the training (fine-tuning) process and can yield good performance even without fine-tuning. 
    {\responserefa{}
    \sys can provide high certified accuracy with low latency when the base model is small, or its patch size is large. Even when defending a large model, we argue the higher latency is an acceptable compromise to the advantage it brings.
    }
\end{tcolorbox}

\section{Discussion}
\label{sec:discussion}


In this section, we further discuss the difference between certified detection and certified recovery. We first discuss a scenario in which the defense of certified detection may be ``bypassed''. This discussion is ignored in previous work and could potentially misguide the readers. Still, we argue that this kind of special case will not influence the practicability of certified detection. Then we discuss one possible direction for improving certified recovery. 

\subsection{A Special Case for Certified Detection}

The certified property in certified recovery and certified detection is different. The former follows the convention defined in digital adversarial defense literature: a model is said to be robust if its inference remains the same in the neighbourhood of one input. The only difference is that the \textit{neighbourhood} is defined as $L_p$ norm in a digital adversarial attack and is defined as a restricted size patch in an adversarial patch attack. To certify such robustness, certified recovery needs to design complicated strategies to ensure the model will yield the same results given only the local features. However, \sys, as well as other certified detection methods, only attempt to detect the presence of abnormal input by making sure that at least one benign voter can control the inference. The philosophy of certified detection can be summarized as follows:

\textit{Truth always rests with the minority, and the minority is always stronger than the majority.}

However, \textbf{What if the minority is wrong}?

\begin{figure}[tbp]
     \centering
     \includegraphics[width=0.9\linewidth]{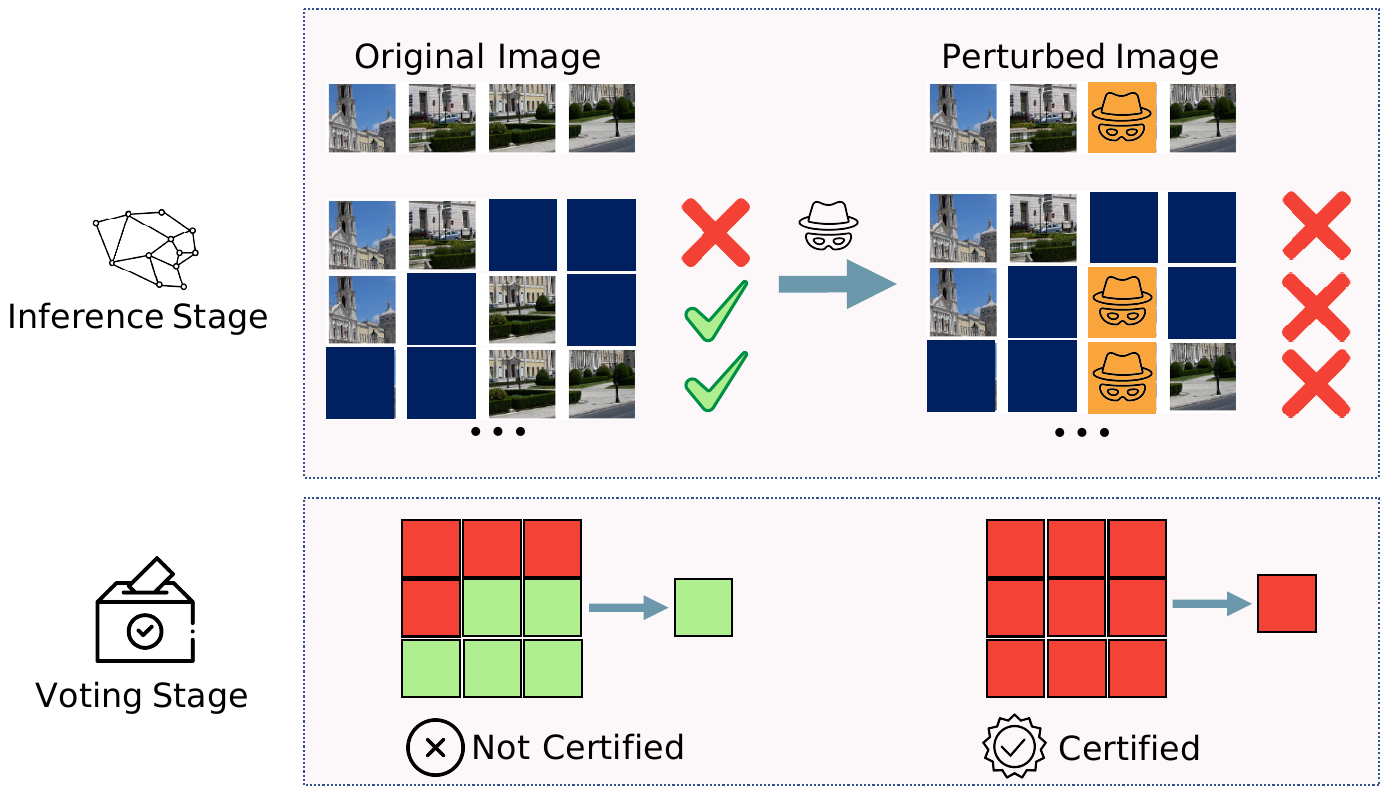}
     \caption{An illustration that certified detection may fail at test time - the method may return `certified' for attacked images.}
     \label{fig:discussion:counter}
     \vspace{-5mm}
 \end{figure}

\begin{figure}[tbp]
    \centering
    \begin{subfigure}[b]{0.3\textwidth}
    \centering
        \includegraphics[width=1.5in]{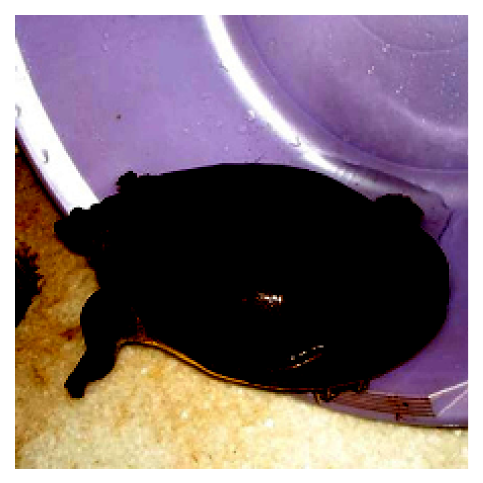}
        \caption{The original image.}
        \label{fig:counter-example:original}
    \end{subfigure}
    ~
    \begin{subfigure}[b]{0.3\textwidth}
    \centering
        \includegraphics[width=1.5in]{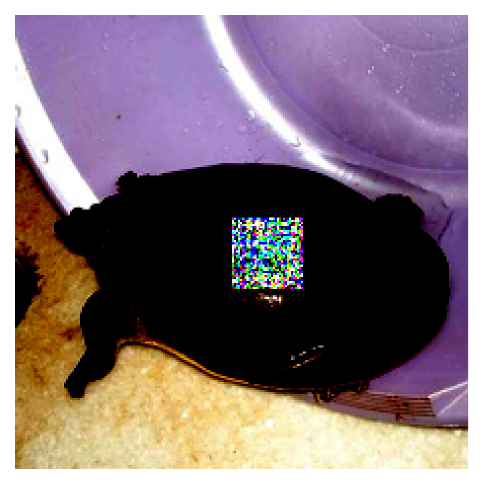}
        \caption{The perturbed image.}
        \label{fig:counter-example:attacked}
    \end{subfigure}
     \caption{A real example that can crack the certified adversarial patch detection. The original image has 69 votes for class 35 (mud turtle) and 53 votes for class 36 (terrapin). The perturbed image has 122 votes for class 36. The ground truth of this image is class 35. As a result, the model is correct but not certified on the clean image and wrong but certified on the perturbed image. However, it is hard to distinguish mud turtles from terrapin even for non-expert humans.}
     \label{fig:counter-example}
\end{figure}

This is when the special case comes and why a test-time robustness guarantee can not be given for \sys and all other certified detection work so far. An illustrative example is shown in Figure~\ref{fig:discussion:counter}. When the majority voting result is the \textit{correct class} while the other voting support one other class (denoted as class B) unanimously, the prediction is correct. However, the certified detection method will return a ``not certified'' result as the voting does not reach a consensus. After the attackers perturb the image, the majority voting can all flip to class B. With the original wrong voting, a consensus is reached, and the certified detection method returns a ``certified'' result by mistake. This subtle difference is ignored in most existing certified detection approaches \cite{xiang2021patchguard++,mccoyd2020minority,han2021scalecert} as the base classifier is assumed to be powerful enough to avoid such cases. For example, in Minority Reports~\cite{mccoyd2020minority}, the authors said ``\textit{... in a benign image, we expect it to be rare for any 3 $\times$ 3 regions in the prediction grid to vote unanimously for an incorrect class}''. However, it is indeed possible for the classifier to misclassify an image unanimously. Even powerful vision transformers can make such mistakes on complex datasets like ImageNet, as we find in experiments. An example of such is shown in Fig~\ref{fig:counter-example}, where the perturbed image is certified as terrapin. 

But before concluding that certified detection could be bypassed, we have to answer another question: \textbf{when will the minority make mistakes?}

The error made by the benign voters is not controlled by any attacker or natural perturbation, as their influence is completely masked out. It is only because the DL models are imperfect and fail to fit the original distribution well. Moreover, this kind of error can be estimated by measuring when the detection method can return a certified and correct result on the clean dataset. This measurement can provide us with a statistical certificate.

Another important fact is that in certified recovery techniques, ``certified robust'' also does not mean ``certified to be correct'', \ie, the images that pass the certification may also be incorrect due to the imprecision of the models. From this perspective, although the robustness property (\ie, model prediction does not change in the neighbourhood of one input) in common sense can be certified, the guarantee for the model to be correct also comes with uncertainty. Actually, we can observe that the $\accInTrust$ of \sys is always higher than the other two SOTA methods in Figure~\ref{fig:evaluation:resize_accuracy}. It is also worth mentioning that in the same experiment, when the ROI is small, there is also a quick drop in accuracy for both certified recovery techniques. 

To sum up, although the certification of \sys (and other certified detection techniques) is a statistical guarantee for a data distribution rather than a test-time guarantee for a certain sample, we believe it is more useful in practice due to its superior performance.


\subsection{A Possible Way to Improve Certified Recovery}

Certified recovery methods aim to use only local features to give reliable predictions. As we have shown in both section~\ref{sec:motivation} and section~\ref{sec:results}, this can significantly influence their certified accuracy. However, additional features such as texture, the material and the object's color may help the classifier predict correctly even if only local features are provided. To confirm this, we additionally conduct an experiment on a toy example. 

MNIST is itself a single-channel image. We add two channels so that those grayscale images become RGB images. We give different digits with different colors and test the performance of PatchSmooth. The certification is done with band size = 2 and the adversarial patch size = 5. The result is shown in table~\ref{tab:discussion:mnist}.

Although this is a toy example, it still demonstrates one possibility to improve the performance of certified recovery methods. In real-world applications, it is possible to improve the robustness of the camera-based autonomous driving system by diversifying the texture of the traffic signs or adding other additional information. Securely adding more information to local patches could be a potential direction for future certified recovery research. 

\begin{table}[]
\caption{Adding additional information can improve the performance of PatchSmooth.}
\label{tab:discussion:mnist}
\begin{tabular}{|c|c|c|}
\hline
                           & Single Channel & Three Channel \\ \hline
Clean Accuracy             & 96.66 \%       & 100 \%        \\ \hline
Certified Image Proportion & 52.84 \%       & 86.03 \%      \\ \hline
$Acc_{in-robust}$          & 99.9 \%         & 100 \%         \\ \hline
\end{tabular}
\end{table}

\section{Threats to Validity}
\label{sec:threats}

We summarize the threats to the validity of our study as follows:

{\noindent\textbf{Internal validity.}} The major threat to validity is whether the proposed method can detect abnormal cases and provide a rigorous guarantee. To mitigate this threat, we carefully develop our mutation strategy and prove its correctness in section~\ref{sec:method}. Through our evaluation, we also find a possible scenario where the detection could fail and give a detailed discussion of the influence in section~\ref{sec:discussion}. Another threat to validity lies in the training/fine-tuning configuration under different experiment settings. These settings can potentially influence the final performance, and it may be unfair to compare different methods if set improperly. To reduce this threat, we refer to the original paper for each method to find the best hyperparameters for training if they are reported. We retrain/fine-tune the model under different experiment settings when it is necessary. 

{\noindent\textbf{Construct validity.}} The construct threats mainly lie in the randomness inherent in our experiments. Specifically, the latency measurement in Section~\ref{sec:results} can be unreliable if only a limited number of instances are tested. To reduce such threats, we conduct our experiments by first resizing all the evaluation instances into the same size and recording the latency of each instance. The final result is given by computing the average of the latency.

{\noindent \textbf{External validity.}} The evaluation dataset could be another threat to generalizability. To mitigate this threat, we evaluate our method on the popular CIFAR-10 and the more complex ImageNet dataset. These two classical evaluation datasets can serve as indicators of how powerful our method is. In addition, to validate \sys can achieve moderate performance when {\responserefa{}ROI} is small, we test \sys on PartImageNet and ImageNet object detection which is more close to real scenarios than the rescaled CIFAR-10.

\section{Conclusion and Future Work}
\label{sec:conclusion}

In this paper, we propose a simple yet effective certified defense against adversarial patches, which utilizes the novel connection between adversarial patches and input patches in Vision Transformers. We demonstrate the effectiveness and practicality of the defense on CIFAR-10, GTSRB, Food-101, and ImageNet, as well as its flexibility to support different sizes of adversarial patches.

\begin{figure}[tbp]
     \centering
     \includegraphics[width=0.75\linewidth]{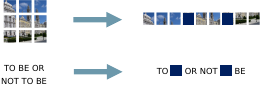}
     \caption{Our masking strategy works on abstract feature level and can be applied to both images and text.} 
     \label{fig:conclusion:mask_demo}
\end{figure}

There are two possible directions for future work. First, we can extend \sys for NLP tasks with some adaptation as our masking strategy works on the abstract feature level, and it does not matter whether the input is image or text, as shown in Fig.~\ref{fig:conclusion:mask_demo}. Another direction is to study the mutation strategy for multiple abnormal patches. The current form of this defense is designed for a single adversarial patch, while other shapes of patches and multiple patches are not handled yet. A problem with multiple patches might be that the number of necessary mutations may be too large, in order to ensure at least one mutation can exclude the adversarial patches. This problem can be solved by using more fine-grained attention masking strategies and/or probabilistic voting mechanisms, which we leave for future work.

\section{Acknowledgement}
\label{sec:acknowledgement}

Yuanchun Li is supported in part by the National Key R\&D Program of China (No.2022YFF0604501) and NSFC (No.62272261).
Yuheng Huang and Lei Ma are supported in part by Canada CIFAR AI Chairs Program, the Natural Sciences and Engineering Research Council of Canada (NSERC No.RGPIN-2021-02549, No.RGPAS-2021-00034, No.DGECR-2021-00019). Lei Ma's research is also supported by JST-Mirai Program Grant No.JPMJMI20B8, JSPS KAKENHI Grant No.JP20H04168, No.JP21H04877. 
We appreciate Michael McCoyd and David Wagner for the enlightening discussion about the advantages and disadvantages of certified detection defense~\cite{mccoyd2020minority}.

\bibliographystyle{ACM-Reference-Format}
\bibliography{reference}





\end{document}